\documentclass{article} % For LaTeX2e
\usepackage{iclr2026_conference,times}

\usepackage{amsmath,amsfonts,bm}

\def\eqref#1{equation~\ref{#1}}
\def\1{\bm{1}}

\def\vl{{\bm{l}}}

\def\vx{{\bm{x}}}
\def\vy{{\bm{y}}}

\DeclareMathAlphabet{\mathsfit}{\encodingdefault}{\sfdefault}{m}{sl}
\SetMathAlphabet{\mathsfit}{bold}{\encodingdefault}{\sfdefault}{bx}{n}

\newcommand{\E}{\mathbb{E}}

\DeclareMathOperator*{\argmax}{arg\,max}

\usepackage{hyperref}
\usepackage{url}
\usepackage{multirow}
\usepackage{booktabs}
\usepackage{xcolor}
\usepackage{xspace}

\usepackage{graphicx}
\usepackage{listings}
\usepackage{booktabs}
\usepackage{multirow}
\usepackage{enumitem}
\usepackage{subcaption}
\usepackage{caption}
\usepackage{framed}
\usepackage{fancybox}
\usepackage{tcolorbox}
\usepackage{wrapfig}

\usepackage{algorithm}
\usepackage{amssymb}
\usepackage{algpseudocode}

\usepackage{booktabs}
\usepackage{multirow} % for \multirow
\usepackage{siunitx}
\colorlet{punct}{red!60!black}
\definecolor{background}{HTML}{EEEEEE}
\definecolor{delim}{RGB}{20,105,176}
\colorlet{numb}{magenta!60!black}

\lstdefinelanguage{json}{
    basicstyle=\scriptsize\ttfamily,
    showstringspaces=false,
    breaklines=true,
    frame=single,
    backgroundcolor=\color{white},
    literate=
     *{0}{{{\color{numb}0}}}{1}
      {1}{{{\color{numb}1}}}{1}
      {2}{{{\color{numb}2}}}{1}
      {3}{{{\color{numb}3}}}{1}
      {4}{{{\color{numb}4}}}{1}
      {5}{{{\color{numb}5}}}{1}
      {6}{{{\color{numb}6}}}{1}
      {7}{{{\color{numb}7}}}{1}
      {8}{{{\color{numb}8}}}{1}
      {9}{{{\color{numb}9}}}{1}
      {:}{{{\color{punct}{:}}}}{1}
      {,}{{{\color{punct}{,}}}}{1}
      {\{}{{{\color{delim}{\{}}}}{1}
      {\}}{{{\color{delim}{\}}}}}{1}
      {[}{{{\color{delim}{[}}}}{1}
      {]}{{{\color{delim}{]}}}}{1},
}

\usepackage{dramatist, etoolbox, enumitem}
\usepackage[framemethod=TikZ]{mdframed}
\makeatletter

\patchcmd{\speaker}{\item[#1\speaksdel]}{\item[\speaksfont#1]}{}{}
\patchcmd{\@character}{\item[#1\speaksdel]}{\item[\speaksfont#1]}{}{}
\makeatother

\title{Fast and Accurate Causal Parallel Decoding using Jacobi Forcing}

\author{
    Lanxiang Hu\thanks{Equal contributions. Part of work was done during Lanxiang's internship at Snowflake.}\textsuperscript{\hspace{1.5mm}}$^{1}$
    \quad
    Siqi Kou\footnotemark[1]\textsuperscript{\hspace{1.5mm}}$^{2}$
    \quad
    Yichao Fu$^{1}$
    \quad
    Samyam Rajbhandari$^{3}$
    \quad
    Tajana Rosing$^{1}$ \\
    \hspace{0.025em}
    \textbf{Yuxiong He}$^{3}$
    \quad
    \textbf{Zhijie Deng}\thanks{Correspondence to Zhijie Deng, Hao Zhang $<$zhijied@sjtu.edu.cn, haozhang@ucsd.edu$>$.}\textsuperscript{\hspace{1.5mm}}$^{2}$
    \quad
    \textbf{Hao Zhang}\footnotemark[2]\textsuperscript{\hspace{1.5mm}}$^{1}$  \\
    \hspace{0.025em}
    $^1$UC San Diego
    \hspace{1pt}
    $^2$ Shanghai Jiao Tong University
    \hspace{1pt}
    $^3$Snowflake
}

\newcommand{\sysname}{Jacobi Forcing Model\xspace}
\iclrfinalcopy % Uncomment for camera-ready version, but NOT for submission.
\begin{document}

\maketitle

\begin{abstract}

Multi-token generation has emerged as a promising paradigm for accelerating transformer-based large model inference. Recent efforts primarily explore diffusion Large Language Models (dLLMs) for parallel decoding to reduce inference latency. To achieve AR-level generation quality, many techniques adapt AR models into dLLMs to enable parallel decoding. However, they suffer from limited speedup compared to AR models due to a \textit{pretrain-to-posttrain mismatch}. Specifically, the masked data distribution in post-training deviates significantly from the real-world data distribution seen during pretraining, and dLLMs rely on bidirectional attention, which conflicts with the causal prior learned during pretraining and hinders the integration of exact KV cache reuse.
% There lacks a parallel decoding method that achieve competitive speedup while maintaining AR-level performance at the same time. 
To address this, we introduce \textit{Jacobi Forcing}, a progressive distillation paradigm where models are trained on their own generated parallel decoding trajectories, smoothly shifting AR models into efficient parallel decoders while preserving their pretrained causal inference property. The models trained under this paradigm, \sysname, achieves $3.8\times$ wall-clock speedup on coding 
% [and math]
benchmarks with minimal loss in performance. Based on \sysname's trajectory characteristics, we introduce multi-block decoding with rejection recycling, which enables up to $4.5\times$ higher token acceptance count per iteration and nearly $4.0\times$ wall-clock speedup, effectively trading additional compute for lower inference latency. Our code is available at \href{https://github.com/hao-ai-lab/JacobiForcing}{https://github.com/hao-ai-lab/JacobiForcing}.

% a progressive trajectory consistency distillation technique that transforms AR models into efficient parallel decoders without altering their pretrained causal inference property.
\end{abstract}

\section{Introduction}
Modern large language models (LLMs), such as GPT-5~\citep{openai2025gpt5}, Gemini-2.5~\citep{google2025gemini2.5}, and Kimi-K2~\citep{team2025kimi}, excel at complex and interactive agentic tasks. Yet, autoregressive (AR) decoding generates tokens sequentially, limiting parallelism and leading to high latency. To address this, recent work explores predicting multiple future tokens natively in transformer-based models without relying on auxiliary draft models. A popular approach is diffusion-based language models (dLLMs), which relax left-to-right generation by modeling the entire sequence jointly and decoding via full-sequence denoising~\citep{nisonoff2024unlocking,schiff2024simpleguidance,khanna2025mercury}. This, in turn, enables highly parallelizable computation. However, open pretrained dLLMs~\citep{ye2025dream7b,zhu2025llada15_vrpo,nie2025large_language_diffusion_models} underperform AR models in generation quality, mainly due to their negative evidence lower bound (NELBO) training objective, a loose bound on AR's negative log-likelihood (NLL) that is proven less efficient~\citep{cheng2025sdar,niescaling,arriola2025block}.

\begin{wrapfigure}{r}{0.48\textwidth}
    \centering
    % \vspace{2.5cm}
    \includegraphics[width=\linewidth]{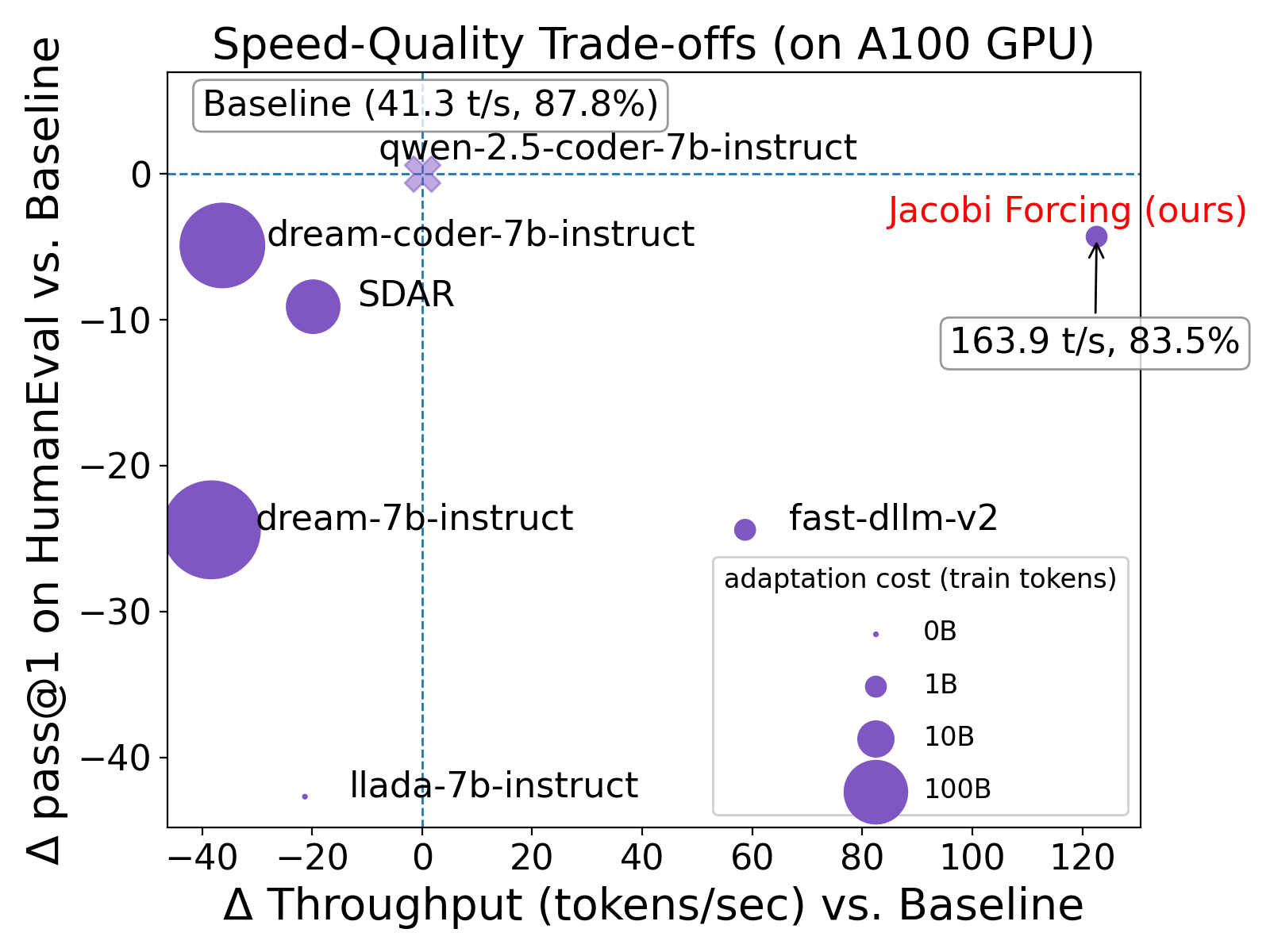}
    \caption{Baseline comparison.}
    % \vspace{-0.25cm}
\label{fig:baselines_comparison}
\end{wrapfigure}

To preserve the generation quality of frontier AR models, the community has adapted high-quality AR models into dLLMs for parallel decoding~\citep{jetastraSDAR, wu2025fastdllmv2efficientblockdiffusion}. Concretely, they perform block-wise perturbations of pretrained data by randomly masking tokens following the dLLMs recipe, and leverage these data to posttrain AR models by modifying the attention mask to enable block-wise bidirectional attention and replacing the training objective from NLL to NELBO. This adaptation delivers limited speedup under quality constraints, primarily due to a significant \textit{pretrain-to-posttrain mismatch}. Specifically, enforcing block-wise bidirectional attention conflicts with the causal prior in pretrained AR models. For instance, SDAR~\cite{cheng2025sdar} suffers substantial quality drops when large block sizes (\textit{e.g.}, 64 or 128) are adopted. Moreover, the masked data distribution during post-training deviates sharply from the natural data distribution seen during pretraining, making the adaptation difficult to learn. Consequently, AR-adapted dLLMs are costly to train as shown in Figure~\ref{fig:baselines_comparison}, and fail to scale speedup reliably with larger block sizes, thereby underutilizing modern AI accelerators whose abundant FLOPs could otherwise be leveraged to decode more future tokens per iteration and further reduce end-to-end latency.

% add figure here to support claims from the above paragraph 

%In this work, we introduce a progressive consistency distillation technique that address the limitation by progressively teaching to predict more tokens within each block and to perform better fast forwarding with increasing block size. We further introduce a noise-aware causal attention that teaches to model to predict correct tokens within each block conditioned on unconverged blocks, and we show it enables more useful future tokens to emerge in each block's trailing tails. We show applying rejection-recycling and multi-block decoding to leverage this model behavior from progressive consistency LLMs (\sysname) for further efficiency improvement.
In this work, we introduce \textit{Jacobi Forcing}, a progressive distillation technique that addresses this \textit{pretrain-to-posttrain mismatch}. It trains AR models on their own generated data without any modifications of causal attention. This is made possible by collecting trajectories using Jacobi Decoding—a widely adopted parallel decoding technique for AR models~\citep{song2021accelerating, santilli2023accelerating}. It first randomly initializes a block of n tokens and feeds it to the AR models to iteratively update it, and eventually, the block converges to the same n tokens generated by AR decoding, forming a trajectory between the randomly initialized point and the converged point. The full sequence is generated block by block. Prior works including CLLM~\citep{kou2024cllms_consistency_large_language_models} and CEED-VLA~\citep{song2025ceed_vla_consistency_vla} design a consistency loss to map any point along the trajectory to the converged point, which in turn teaches AR models to predict multiple correct tokens in one iteration simultaneously. However, they face a similar limitation as AR-adapted dLLMs: as block size increases, the number of tokens correctly decoded per iteration remains essentially constant. \textit{Jacobi Forcing} addresses this by introducing a noise-aware causal attention that teaches the model to predict the converged point within each block conditioned on previous unconverged blocks, and we show it enables more useful future tokens to emerge in each block's trailing tails. Furthermore, \textit{Jacobi Forcing} repeats this distillation procedure for the trained model and involves more noisy data with a larger block size for progressive distillation. 
%We dub models trained under \textit{Jacobi Forcing} as \sysname.

%Moreover, 
We observe \sysname has a stronger capability of generating correct future tokens conditioning on noisy context, consistent with our training objective. To better utilize this characteristic, we design a rejection-recycling and multi-block decoding algorithm for further inference optimization. Rejection recycling reuses high-quality consecutive tokens discarded from past Jacobi iterations to generate candidate token sequences, enabling the decoding of more accurate tokens via verifying multiple branches in a single iteration. Multi-block decoding maintains and refines multiple blocks simultaneously, where correct tokens are decoded in subsequent blocks even when preceding blocks remain unconverged for further speedup.

Experiments show \sysname can serve as very efficient parallel decoders with up to $3.8\times$ improvement in generation speed across coding and math benchmarks. It also effectively generates higher quality draft n-grams from future tokens within each block, as observed in Section~\ref{sec:exp}. Using rejection-recycling and multi-block decoding makes use of future n-grams and further boost speedup to around $4\times$. 

In summary, key contributions of this paper includes:
\begin{itemize}[left=10pt, itemsep=0pt]
\item We introduce \textit{Jacobi Forcing} to train AR models as fast parallel decoders, \sysname, with up to $3.8\times$ generation speedup.
\item We empirically observe and qualitatively verify \sysname has both higher fast-forwarded token count and a useful n-gram count in comparison with baseline models. 
\item We propose rejection-recycling and multi-block decoding to make use of higher quality draft n-grams from future tokens within each block, and apply them to \sysname boost generation speed nearly up to $4.0\times$ across various benchmarks.
\end{itemize}

\section{Preliminary}
This section reviews the basics of Jacobi decoding and consistency distillation training to accelerate Jacobi decoding of AR models.

\subsection{Jacobi Decoding} 
\label{sec:jacobi_decoding}
Given a prompt $\vx$ and a pre-trained LLM $p_\theta(\cdot | \vx)$ parametrized by $\theta$, the standard AR decoding under the greedy strategy produces a response sequentially as follows:
\begin{equation}
\label{eq:ar_decoding}
y_i = \underset{y}{\arg\max}  \;  p_\theta(y \mid \vy_{<i}, \vx), \quad \text{for } i = 1, \dots, n,
\end{equation}
where $\vy_{<i} = \{y_1, \dots, y_{i-1}\}$. This process requires $n$ forward passes of the LLM to generate $n$ tokens $\vy_{\leq n}$. The inherently sequential nature of AR decoding limits practical efficiency when generating long sequences. Jacobi decoding~\citep{song2021accelerating,santilli2023accelerating} addresses this bottleneck by reformulating token generation as solving a system of nonlinear equations:
\begin{equation}
 f(y_i, \vy_{<i}, \vx) = 0, \quad \text{for } i = 1, \dots, n,
 \label{eq:jacobi_eq}
\end{equation}
where $f(y_i, \vy_{<i}, \vx):= y_i-{\mathrm{arg\,max}_y}\ p_\theta(y | \vy_{<i}, \vx)$. This system can be solved in parallel using Jacobi fixed-point iteration~\citep{ortega2000iterative}. Starting from a randomly initialized $n$-token sequence $\displaystyle {\vy^{(0)}=\{y_{1}^{(0)},  \ldots, y_{n}^{(0)} \} }$, the update at each iteration $j$ is:
\begin{equation}
\label{eq:jacobi_decoding}
\begin{aligned}
\begin{cases}
y_{1}^{(j+1)} &= \underset{y}{\mathrm{arg\,max}}\ p_\theta(y | \vx) \\
y_{2}^{(j+1)} &= \underset{y}{\mathrm{arg\,max}}\ p_\theta(y | \vy_{1}^{(j)}, \vx) \\
& \vdots \\
y_{n}^{(j+1)} &= \underset{y}{\mathrm{arg\,max}}\ p_\theta(y | \vy_{<n}^{(j)}, \vx).
\end{cases}
\end{aligned}
\end{equation}
Notably, for LLM, the above $n$ maximization problems can be solved in parallel by using a causal attention mask, i.e., only one forward pass of the LLM is required to obtain $\vy^{(j+1)}$ based on $\vy^{(j)}$. 
The iteration exits at some $k$ such that $\vy^{(k)}=\vy^{(k-1)}$ and we define $\vy^*:=\vy^{(k)}$ as the fixed point. Let $\mathcal{J} := \{\vy^{(0)}, \dots, \vy^{(k)}\}$ denote the Jacobi trajectory. It can be proven that $\vy^*$ is identical to AR decoding under greedy strategy~\citep{song2021accelerating}. 

To generate a long response $\vl$ of length $L \gg n$, Jacobi decoding is applied sequentially over blocks of size $n$ until the \texttt{<eos>} token appears in a fixed point. Let $\vy_{B_i}^*$ denote the fixed point obtained for the $i$-th block. The full output $\vl$ is then constructed by concatenating fixed points from consecutive blocks:
\begin{equation}
    \vl= [\vy_{B_1}^{*}, \dots, \vy_{B_N}^{*}], 
\end{equation}
where $N= \lceil \frac{L}{n} \rceil$ denotes the number of blocks generated before termination.

\subsection{Consistency Distillation} 
Despite the promise, Jacobi decoding achieves little speedup over standard AR decoding~\citep{santilli2023accelerating,fu2024lookahead}, as it rarely predicts more than one correct\footnote{By correctness, we mean alignment with the AR decoding result under a greedy sampling strategy.} token within one fixed-point iteration. To address this, recent works such as CLLMs~\citep{kou2024cllms_consistency_large_language_models} propose consistency distillation, a training approach designed to accelerate convergence to the fixed point from arbitrary states on a Jacobi trajectory. The key idea is to introduce a consistency loss that encourages an LLM $p_\theta(\cdot|\vx)$ to predict multiple tokens simultaneously:
\begin{equation}
\label{eq:cllm1_consistency_loss}
    \begin{aligned}
    \mathcal{L}_{\text{c}} &= \E_{ i\sim \mathcal{U}\{1, \ldots, N\}, \vy_{B_i} \sim \mathcal{J}_i} \Big[D_{\text{KL}}\left( p_{\theta^{-}}(\vy_{B_i}^*|\vx, \vy_{B_1}^*, \ldots, \vy_{B_{i-1}}^*) || p_{\theta}(\vy_{B_i}|\vx, \vy_{B_1}^*, \ldots, \vy_{B_{i-1}}^*) \right) \Big],
    \end{aligned}
\end{equation}
where $\theta^{-} = \text{stopgrad}(\theta)$ and $D_{\text{KL}}$ denotes the KL divergence aggregated across the $n$ tokens in a block. Here, $i\sim \mathcal{U}\{1, \ldots, N\}$ denotes sampling a block index uniformly at random, and $\vy_{B_i} \sim \mathcal{J}_i$ denotes randomly sampling from the Jacobi trajectory of the $i$-th block.

CLLMs build upon this idea by first collecting Jacobi trajectories, obtained by running Jacobi decoding with $p_\theta$ on a set of prompts. The model is then trained with a joint objective that combines the consistency loss in Eq.~\ref{eq:cllm1_consistency_loss} with the standard AR loss, achieving up to a $2\times$ speedup over AR decoding while maintaining quality. Similar training objectives have also been adopted for inference acceleration in other domains, such as action prediction in VLA models~\citep{song2025ceed_vla_consistency_vla}.

\section{Methodology}
In this section, we first discuss the training challenges of consistency distillation with larger block sizes $n$, and then present \textit{Jacobi Forcing}, a progressive consistency distillation method designed to mitigate this bottleneck, and denote LLMs trained under this paradigm as \sysname. Furthermore, by observing \sysname's trajectories under vanilla Jacobi decoding, we introduce rejection-recycling and multi-block decoding strategies to improve its efficiency.

\subsection{Jacobi Forcing}
\label{sec:cllm2}

\textbf{Progressive Noise Schedule.}
%Sequence packing. Specify the attention mask.
In Jacobi decoding, we maintain strict causality within each block, where each token is updated in accordance with Eq.~\ref{eq:jacobi_decoding}.
Consider the $i$-th block $\vy^{\left(j\right)}_{B_i}$ of size $n$ is been decoded at some iteration step $j$. Assume the first $c-1$ tokens have been accepted, and we denote 
%$y_f \in \vy_{>j+1} = \{y_{j+2}, ..., y_n\}, j+ 2 \leq f \leq n$ 
$y_f$ as the future token as shown in Eq.~\ref{eq:future_token_decoding}.
%where $\vy_{i<c}$ is the clean context, $\vy'_{c<i<f}$ is the noisy context with unconverged tokens. Predicting useful the future tokens within each block with a causal mask is hard because each $y_f$ is now conditioned on at least one unconverged noisy token $y_{j+1}$ according to Eq.~\ref{eq:ar_decoding}. 
\begin{equation}
%y_f = \underset{y}{\arg\max}  \;  p\big(y \mid \vy'_{c<i<f}, \vx_c \big), \quad \text{for } 1 \leq c < i < f \leq n,
%= p\big(y_f \mid y_1,\, y_2,\, \dots,\, y_{f-1},\, \underbrace{y_{j+1}, y_{j+2},\, \dots,\, y_{f-1}}_{\mathrm{noisy}}\big)
y_f = \argmax_{y} \; p\!\left(y \mid \vx_c, \; \vy'_{c:f-1} \right),
\quad \text{for } f=c+1, \dots, n,
%\forall  \, c < f \leq n,
\label{eq:future_token_decoding}
\end{equation}
where $\vx_c = \left[ \vx, \vy_{<c} \right]$ is the clean context, $\vy'_{c:f-1}$ is the noisy\footnote{By noisy, we refer to tokens in the non-converged point along the Jacobi trajectory that that differ from those in the fixed point at the same positions.} context. While the training objective in Eq.~\ref{eq:cllm1_consistency_loss} is designed to optimize correct token prediction in this setting, it's observed from \cite{kou2024cllms_consistency_large_language_models} that predicting $y_f$ is hard when it's conditioned on a long noisy context $\vy'_{c:f-1}$ under large block sizes (e.g., $n=256$). 

To address this challenge, we instead split a large block into smaller blocks (e.g., $n=16$) with noise ratios determined by a predefined schedule $\{t_1, \dots, t_N\}$. Each $t_i$ denotes the fraction of noisy tokens in a block. The noise schedule follows a cyclic strategy with window size $w$, where the noise ratio linearly increases from 0 to 1 within each window, i.e.,
\begin{equation}
    \label{eq:noise_schedule}
    W=\left\{0, \frac{1}{w}, \dots, \frac{w-1}{w}\right\}, \quad t_i = W[j], \quad j=i \bmod w.
\end{equation}
This progressive schedule ensures that each block retains a partially clean context, thereby shortening noisy tokens dependencies. In particular, it reduces the longest span of consecutive noisy inputs for any prediction from $O(nN)$ assuming $t_i=1$ for all blocks using a random schedule to $O\!\left(\lceil tn \rceil\right)$ using a progressive schedule, which facilitates learning. Empirically, we find this progressive schedule to be more effective than a purely random noise schedule (Table~\ref{tab:noise_schedule}). 

\begin{figure}[t]
    \centering
    % -------- left image ----------------------------------------
    \begin{minipage}[c]{0.49\linewidth}
        \centering
        \vspace{-1cm}
        \includegraphics[width=\linewidth]{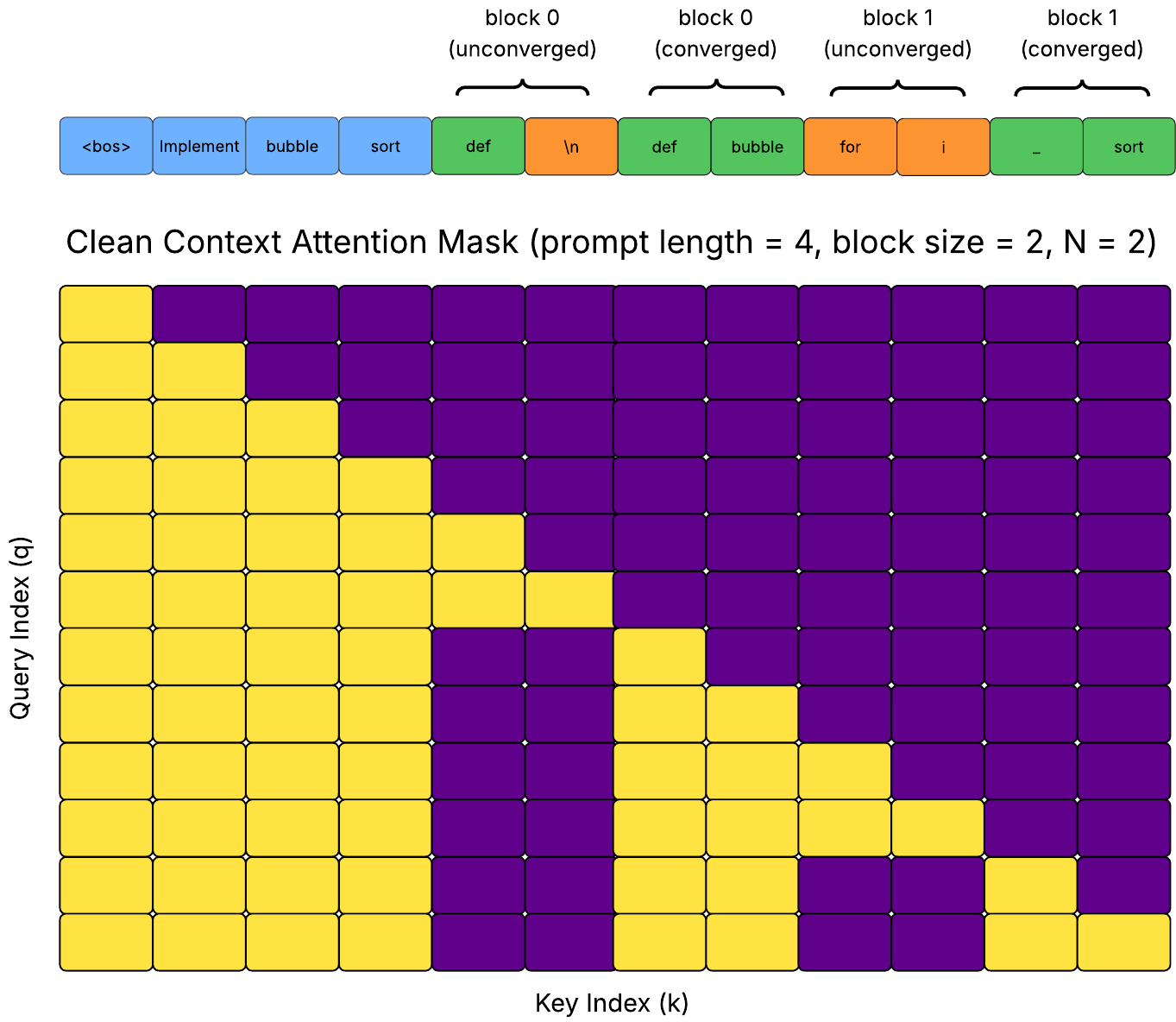}
        \subcaption{clean-context conditioned causal mask.}
        \label{fig:clean_context_causal_mask}
    \end{minipage}
    % -------- right image ---------------------------------------
    \begin{minipage}[c]{0.49\linewidth}
        \centering
        \vspace{-1cm}
        \includegraphics[width=\linewidth]{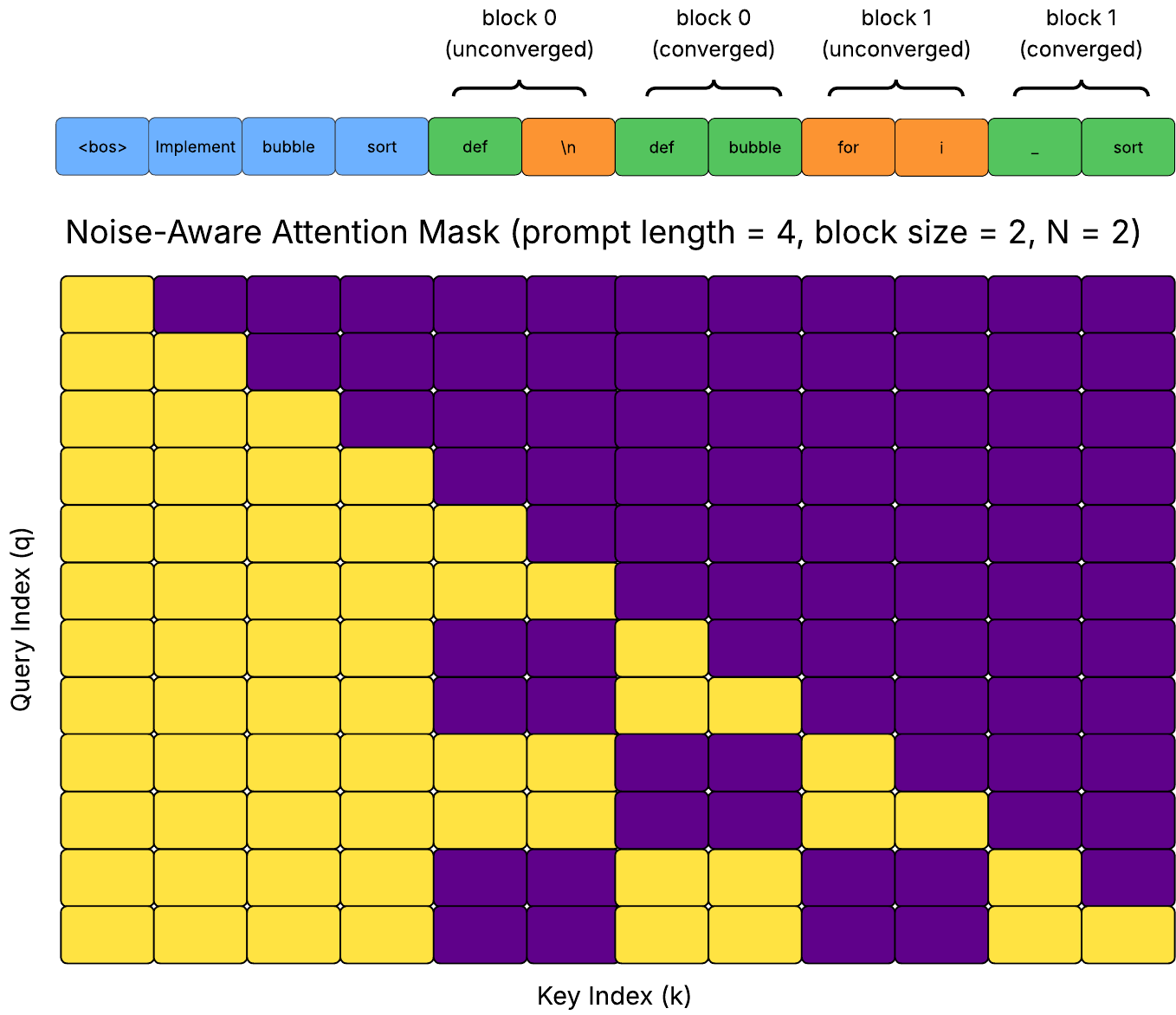}
        \subcaption{noisy-context conditioned causal mask.}
        \label{fig:noise_aware_causal_mask}
    \end{minipage}
    \hfill
    % ------------------------------------------------------------
    \caption{Sequence packing with two attention mask implementations, both allow logits from clean blocks and noisy blocks to be generated with single forward pass to calculate the progressive consistency loss and AR loss in Eq.~\ref{eq:cllm2_total_loss}.}
    \label{fig:sequence_packing_and_mask}
\end{figure}

\textbf{Progressive Distillation Loss.} Let $\vy_{b_i}^{t_i}$ denote the point along the $i$-th block Jacobi trajectory with several noisy tokens closest to $\lceil t_i n \rceil$. The training objective is to predict tokens correctly within each block, aggregating losses across blocks to reduce gradient variance and stabilize optimization. Accordingly, we introduce a new loss term, \emph{progressive consistency loss}, which optimizes $p_\theta$ under the progressive noise schedule in Eq.~\ref{eq:noise_schedule}:

\begin{equation}
\label{eq:cllm2_consistency_loss}
\begin{aligned}
\displaystyle \mathcal{L}_{\text{pc}} 
&= \dfrac{1}{N} \sum_{i=1}^N 
D_{\text{KL}} \big( p_{\theta^{-}}( \cdot \mid \vx, \vy_{B_1}^*, \ldots, \vy_{B_{i-1}}^*) \big\Vert\ p_{\theta}( \cdot \mid \vx, \vy_{B_1}^{t_{1}}, \ldots, \vy_{B_{i-1}}^{t_{i-1}}) \big).
\end{aligned}
\end{equation}

\textbf{AR Loss.} \citet{kou2024cllms_consistency_large_language_models} notes that using only the consistency loss (Eq.~\ref{eq:cllm1_consistency_loss}) must be supplemented with an AR loss to maintain generation quality. Our preliminary experiments show that using only the consistency objective (Eq.~\ref{eq:cllm2_consistency_loss}) produces the same effect. This motivates our inclusion of a conventional AR loss term in the final training objective to safeguard output quality:
\begin{equation}
\label{eq:cllm2_total_loss}
\begin{aligned}
\displaystyle \mathcal{L}(\theta) = \mathcal{L}_{\text{pc}} + w \mathcal{L}_{\text{AR}}
\end{aligned}
\end{equation}
where $w$ is a tunable weight that balances the two learning objectives.

\textbf{Noise-aware Causal Attention.} In CLLM, loss from each training step is computed based on KL divergance from one block instance in Eq.~\ref{eq:cllm1_consistency_loss}. This learning objective is to train correct token prediction in the setting where there is only a big block (Eq.~\ref{eq:future_token_decoding}). 
%To enable the decoding paradigm in Eq.~\ref{eq:noise_conditioned_decoding_within_window} as well as the learning objective in Eq.~\ref{eq:cllm2_consistency_loss}, 
Moreover, in both Eq.~\ref{eq:cllm1_consistency_loss} and Eq.~\ref{eq:cllm2_consistency_loss}, the loss term computation involves two forward passes using a conventional causal mask since each involves a distinction sequence. As a result, it requires $O(2N)$ forward passes to compute all loss terms in Eq.~\ref{eq:cllm2_consistency_loss} and $O(N)$ backward passes to compute gradients, resulting in low training efficiency. We reduce the number of forward and backward passes from $O(N)$ to $O(1)$ by introducing a sequence packing technique and a block-wise sparse attention mask. We illustrate the sequence packing that interleaves $\vy_{b_i}^{t_{i}}$ and $\vy_{b_i}^*$ for the entire complete sequence in Figure~\ref{fig:noise_aware_causal_mask} for $\mathcal{L}_{pc}$ computation, in contrast with conditioning each unconverged $y_{b_s}$ only on clean tokens for consistency distillation with $\mathcal{L}_{c}$ in Figure~\ref{fig:clean_context_causal_mask}.

%\lx{add illustrations: cllm1 attention with clean context vs. cllm2 attention conditioned on noise}

\textbf{Progressive Distillation for Larger Block Sizes.} In training \sysname on trajectories from the original AR model, we find that speedup scales with training steps and saturates at large step counts, likely due to significant data distribution shifts from extensively trained models. To break this ceiling, we collect an additional round of Jacobi trajectories with \emph{progressively larger block sizes} from the \sysname empowered with multi-token prediction capability and further train it on newly generated trajectories. This yields a further 20\% speedup with only minor performance degradation. Detailed training configurations are in Section~\ref{sec:exp_settings}.
%We present the algorithm of progressive distillation with increasing block sizes in Algorithm~\ref{alg:progressive_block_size} and we present more experiment details on how speedup and model performance change in Section~\ref{sec:exp_results}.

%\todo{add pseudo code to illustrate? maybe visualization as well? @Siqi}

\subsection{Inference Optimization} 
\label{sec:cllm2_inference}

\begin{figure}[t]
    \centering
    \includegraphics[width=1.0\linewidth]{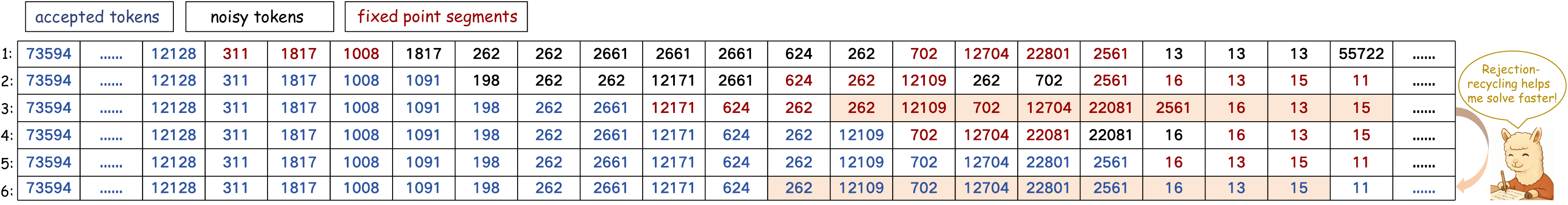}
    \caption{Visualization of \sysname's trajectory under vanilla Jacobi decoding. The figure shows a partial segment of the trajectory. Blue tokens denote accepted tokens that match the fixed point at their positions. Black tokens denote unconverged noisy tokens, and we highlight them in red if more than three consecutive tokens match the fixed point regardless of position.}
    \label{fig:trajectory_vis}
    \vspace{-0.25cm}
\end{figure}

%\textbf{Trajectory behavior in vanilla Jacobi decoding.} 
\textbf{Behavior of \sysname.}
\sysname is trained to have a stronger capability of generating correct future tokens conditioning on noisy tokens. Qualitative analysis in Figure~\ref{fig:trajectory_vis} illustrates that it indeed brings the quality improvement: fixed-point segments emerge within the noisy tokens of the unconverged point. Furthermore, these segments progressively extend (e.g., the number of red tokens increases from point 1 to point 2 in Figure~\ref{fig:trajectory_vis}), even under noisy context, consistent with our training patterns. In this section, we focus on how to translating this qualitative observation of draft quality improvement into qualitative speedup.

\textbf{Rejection Recycling.} Prior work has shown that n-grams produced during Jacobi iterations can be verified in parallel and reused in subsequent iterations~\citep{fu2024lookahead}. As illustrated in Figure~\ref{fig:trajectory_vis}, such n-gram sizes could be large in \sysname. If correctly verified, many tokens can be fast-forwarded in one iteration. In particular, we initialize a fixed-size n-gram pool constructed from noisy token sequences observed at unconverged points during Jacobi decoding. During decoding, if the pool contains an n-gram whose first token matches the last accepted token of the current point, we extend this token by concatenating it with its subsequent tokens to form new candidates. These candidates are then verified in parallel by appending them along the batch dimension. At each iteration, we select the candidate that yields the largest number of newly accepted tokens. For instance, this strategy enables skipping from point 3 to point 5 in Figure~\ref{fig:trajectory_vis}, as the fixed-point segments in point 3 yield higher-quality candidates.

\begin{wrapfigure}{r}{0.56\textwidth}
    \centering
    % \vspace{-0.5cm}
    \includegraphics[width=\linewidth]{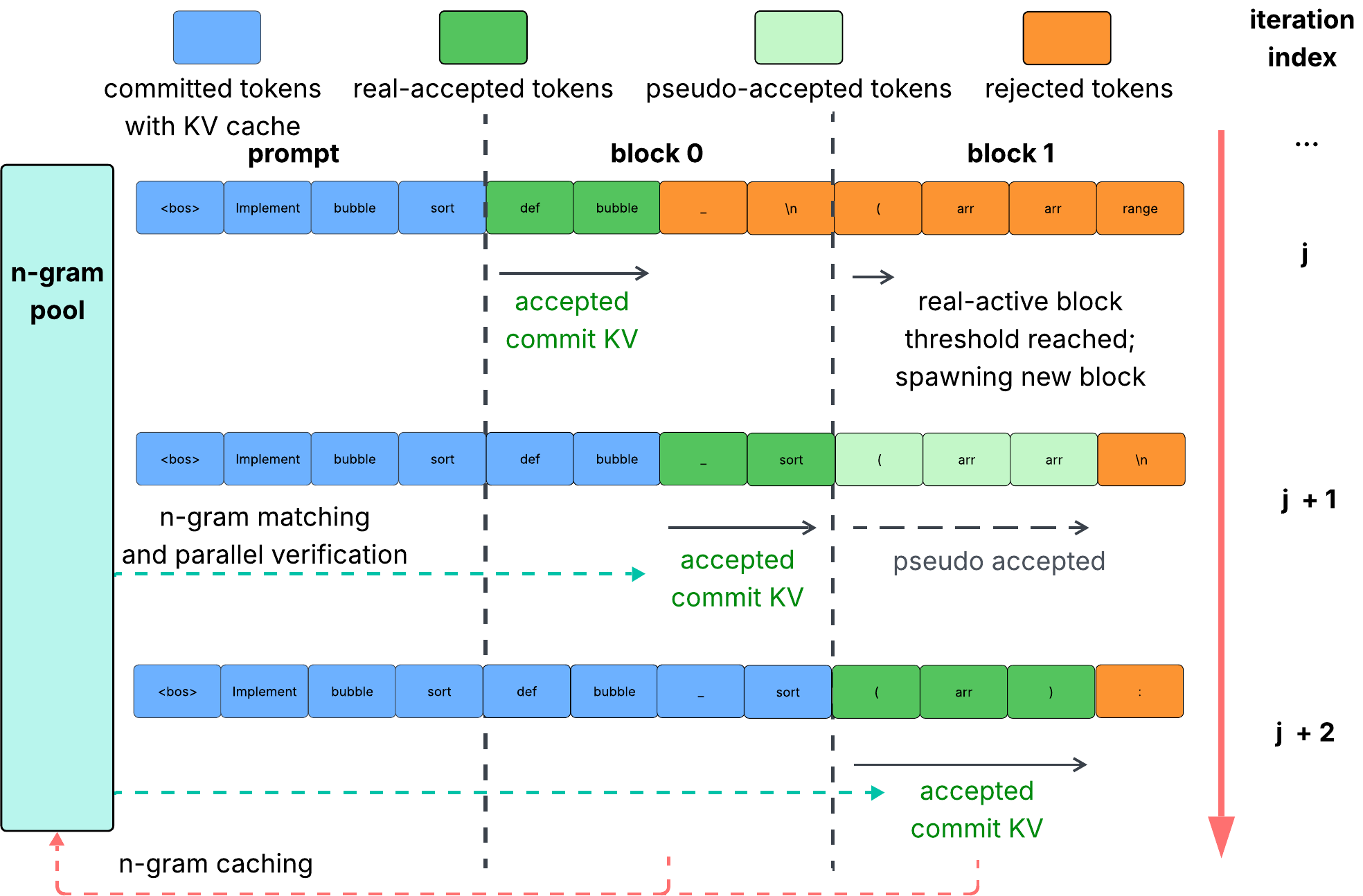}
    \caption{An example of multiblock decoding with rejection recycling at prompt length = 4, block size = 4, $r=0.5$, $K = 2$.}
    \vspace{-0.25cm}
    \label{fig:mutiblock_and_rejection_recycling}
\end{wrapfigure}

%\begin{figure}
%    \centering
%    \includegraphics[width=\linewidth]{figures/multiblock_rejection_recycling_v4.pdf}
%    \caption{An example of multiblock decoding with rejection recycling at prompt length = 4, block size = 4, $r=0.5$, $K = 2$.}
%    \label{fig:trajectory_vis}
%\end{figure}

\textbf{Multi-block Decoding.} In addition to high-quality n-grams in the draft, we also observe the increasing number of stationary tokens, which are correctly predicted with preceding noisy tokens and remain unaltered through subsequent iterations. Together they yield higher quality drafts. To make use of the property, we introduce \emph{multi-block decoding}, a new decoding paradigm that maintains and refines up to $K$ blocks simultaneously. It marks the block closest to the effective KV cache boundary as the \emph{real-active} block and all the other $K-1$ blocks as \emph{pseudo-active} blocks. Only tokens within the real-active block are accepted and committed to KV cache. Tokens in pseudo-active blocks are only pseudo-accepted, conditioning on prior blocks; once converged, pseudo-active blocks will wait until they are promoted as the real active block, where all tokens will be verified again, but now with a higher-quality draft. A detailed description is provided in Algorithm~\ref{alg:multiblock} (with rejection recycling) in Appendix~\ref{appendix:algorithm} and with an example in Figure~\ref{fig:mutiblock_and_rejection_recycling}. Note that both rejection recycling and multi-block decoding are lossless as they employ greedy rejection sampling for token acceptance in the real-active block~\citep{leviathan2022speculative_decoding}.

% \input{ICLR26-CLLM2/codes/multiblock_lookahead_decoding}

%\todo{add algorithm block for rejection cycling + multiblock.}

%\lx{add visualization on how the algorithm works}
\section{Experiments}
\label{sec:exp}

\subsection{Evaluation Settings}
\label{sec:exp_settings}

\textbf{Models and Datasets.} We evaluate \sysname across coding 
%and mathematical benchmarks. 
benchmark.
For coding benchmarks, we train Qwen2.5-Coder-Insutrct~\citep{hui2024qwen2coder} on OpenCodeInstruct~\citep{ahmad2025opencodeinstruct} and test on the HumanEval~\citep{chen2021evaluating}, MBPP~\citep{austin2021program}. 
%For mathematical tasks, we train Qwen3-4B-Instruct-2507~\citep{yang2025qwen3_technical_report, qwen3_4b_instruct_2507} on the math split of Openthought2~\citep{guha2025openthoughtsdatarecipesreasoning} and test on GSM8K~\citep{cobbe2021gsm8k}, and MATH~\citep{hendrycks2021math}.
%and AMC23~\citep{maa2023amc}. 
On OpenCodeInstruct, we curate question instances that come with generations that pass all unit tests, from where we use 450k prompts for trajectory generation and training. 
%On Openthought2, only mathematical prompts are considered, from where we use around 400k prompts for trajectory generation and training. 
For mathematical tasks, we train Qwen2.5-Math-7B-Instruct~\citep{yang2024qwen25mathtechnicalreportmathematical} on the math split of Openthought2~\citep{guha2025openthoughtsdatarecipesreasoning} and test on GSM8K~\citep{cobbe2021gsm8k}, and MATH~\citep{hendrycks2021math}.
On Openthought2, only mathematical prompts are considered, from where we apply the same training settings for trajectory generation and training. 

\textbf{Training Settings}. All training and inference are conducted on instances equipped with 8x NVIDIA A100-80GB GPUs and 8x NVIDIA H200 GPUs. All models are trained with a learning rate of $10^{-6}$, a batch size of 4, and a max new sequence length of 2048. For \sysname, we adopt a linear progressive noise schedule, initial block size at 16, window size at 16, and train for 10k steps, and a second round of training with block size at 32, window size at 8, and train for another 10k steps. Ablation studies on parameter choices are presented in Section~\ref{sec:exp_ablation}.

\textbf{Baselines.} Our main objective in this section is to compare performance and efficiency between diffusion-based parallel decoders and the AR-based parallel decoder, \sysname. The dLLM baselines also have the capability of generating a single block of tokens or multiple consecutive blocks of tokens together. Specifically, we compare \sysname with state-of-the-art (SOTA) dLLMs including LLaDA-7B~\citep{nie2025llada}, Dream-7B~\citep{ye2025dream7b}, fast-dLLM~\citep{wu2025fast_dllm} and D2F~\citep{wang2025diffusion_forcing}. We also compare \sysname with AR-based parallel decoder, including vanilla Jacobi decoding~\citep{santilli2023accelerating} and CLLM~\citep{kou2024cllms_consistency_large_language_models}. In this work, we do not focus on speculative decoding methods, because the models themselves don't serve as parallel decoders without supplemental architecture modifications (e.g. via additional heads)~\citep{cai2024medusa, li2024eagle, li2024eagle2, li2025eagle3} or separate draft models~\citep{leviathan2022speculative_decoding, liu2024online_speculative_decoding}.
In addition, to situate \sysname among broader AR-acceleration techniques, we present in the Appendix~\ref{appendix:additional_comparisons} a complementary comparison with speculative decoding and consistency-distilled baselines.

\subsection{Results}
\label{sec:exp_results}

\begin{table}[t]
\centering
% \vspace{-0.5cm}
\small
\setlength{\tabcolsep}{6pt}
\caption{Performance and efficiency on coding benchmarks, HumanEval and MBPP, grouped by decoding family. For AR-based models, all methods adopt Qwen2.5-Coder-7B-Instruct. For \sysname, MR stands for employing the multi-block and rejection-recycling decoding algorithm introduced in Algorithm~\ref{alg:multiblock}. DC stands for using bi-directional dual cache from fast-dLLM. For both Fast-dLLM and D2F, we choose the Dream-7B as it's significantly faster with similar or better performance than LLaDA-7B. For CLLM*, we follow mostly the same recipe in CLLM but with new sequence packing technique (without progressive training on larger block sizes). The speedup ratio is relative to the AR baseline. 
%Same naming convention is adopted in Table~\ref{tab:math_benchmarks}.
}
\resizebox{\linewidth}{!}{%
\begin{tabular}{lll c c c c}
\toprule
Benchmark & Family & Method & {TPF $\uparrow$} & {TPS $\uparrow$} & {Speedup $\uparrow$} & {Accuracy $\uparrow$}  \\
\midrule
\multirow{9}{*}{\textbf{HumanEval}} 
& \multirow{5}{*}{\textit{AR-based}}    & AR          & 1.00 &  41.3    &   1.00$\times$    &  87.8  \\
&                                        & Jacobi    & 1.03  &   39.9   & 0.97$\times$      & 87.8  \\
&                                      & CLLM*        &  2.68   &   103.3   & 2.50$\times$   &  87.8 \\
&                                         & \sysname     &  4.01 &  159.5  &  3.86$\times$   & 83.5  \\
&                                        & \sysname (MR)  &  \textbf{4.09}  &  \textbf{163.9} & \textbf{3.97$\times$}  &  83.5  \\
\cmidrule(l){2-7}
& \multirow{4}{*}{\textit{Diffusion-based}} 
& LLaDA-Instruct            &   1.00       & 2.8     & 0.07$\times$  & 36.0      \\
&           & Dream-Base     &   1.00    & 20.2    & 0.49$\times$  &  54.3 \\
%&                        & Fast-dLLM (DC)       & 19.2  & 5.2   & 35.4 \\
&      & Fast-dLLM (DC)    &    1.80     & 60.0  &  1.45$\times$  & 53.0   \\
%&                        & D2F-LLADA          & 81.6 & 1.6 & 40.2 \\
&     & D2F              &     2.50     & 73.2   &  1.77$\times$ &  54.3  \\
\midrule
\multirow{9}{*}{\textbf{MBPP}} 
& \multirow{5}{*}{\textit{AR-based}}   & AR        &    1.00     &  43.1  &  1.00$\times$   & 74.3 \\
&                                     & Jacobi    &    1.01     &     42.4 &  0.98$\times$   &   74.3  \\
&                                     & CLLM*    &     2.10      & 80.1    &   1.94$\times$  &  71.4   \\
&                                   & \sysname   &    2.74      &  110.7  &  2.57$\times$   & 70.4  \\
&                               & \sysname (MR)  &     \textbf{2.84}     &  \textbf{113.0}   &   \textbf{2.62$\times$}  & 70.4  \\
\cmidrule(l){2-7}
& \multirow{4}{*}{\textit{Diffusion-based}} 
& LLaDA-Instruct           &     1.00    & 0.9    &  0.02$\times$ & 39.0  \\
&       & Dream-Base       &    1.00    & 10.4   &  0.24$\times$  & 56.2 \\
&       & Fast-dLLM (DC)   &     1.90   & 73.2   &  1.70$\times$  & 51.0  \\
&        & D2F            &      2.30  & 105.0 &  2.44$\times$ & 55.2 \\
\bottomrule
\end{tabular}
}%
\label{tab:coding_benchmarks}
\end{table}

\textbf{Performance.} The performance metrics are the greedy generations’ 
%problem solve rate (test@1) on GSM8K, MATH, 
%and AMC23 
%as well as and 
strict accuracy (pass@1) on HumanEval and MBPP. Table~\ref{tab:coding_benchmarks} 
%and Table~\ref{tab:math_benchmarks} 
compares \sysname with both dLLMs and Jacobi decoding baselines.
On A100 GPUs, our results show that on both benchmarks, \sysname consistently achieves competitive accuracy with a much better speedup at the same parameter scale.
In particular, for structured generations like Python coding, \sysname achieves $3.6\times$ speedup in comparison with the AR baseline, $53.3\sim7.4\times$ speedup comparing to dLLM baselines, and 2.0$\times$ comparing to optimized dLLM baselines including Fast-dLLM and D2F with techniques like adding block-wise KV cache, bidirectional KV cache and pipelined parallel decoding. For speedup evaluation, we run all evaluations with a block size of 128 except for \sysname (MR) since MR takes extra FLOPs for multiblock decoding and parallel verification.

\begin{table}[t]
\centering
\vspace{-0.2cm}
\small
\setlength{\tabcolsep}{6pt}
\caption{Performance and efficiency on math benchmarks, GSM8K and MATH, grouped by decoding family. For AR-based models, all methods adopt Qwen2.5-Math-7B-Instruct.}
%\begin{tabular}{lll S[table-format=2.1] S[table-format=2.1] S[table-format=2.1] c}
\resizebox{\linewidth}{!}{%
\begin{tabular}{lll c c c c}
\toprule
Benchmark & Family & Method & {TPF $\uparrow$}  & {TPS $\uparrow$} & {Speedup $\uparrow$} & {Solve Rate $\uparrow$}  \\
\midrule
\multirow{9}{*}{\textbf{GSM8K}} 
& \multirow{5}{*}{\textit{AR-based} } & AR        &   1.00    &  41.8 &  1.00$\times$ &  92.4 \\
&                                    & Jacobi     &   1.05    &  42.2  &  1.02$\times$ & 92.4 \\
&                                      & CLLM*    &    2.25    &  86.8 & 2.08$\times$  &  92.2   \\
&                                   & \sysname    &   3.72    &  146.1 &  3.50$\times$ &  91.4 \\
&                                 & \sysname (MR)  &   \textbf{4.04}    & \textbf{154.9}  & \textbf{3.71$\times$}  &  91.4 \\
\cmidrule(l){2-7}
& \multirow{4}{*}{\textit{Diffusion-based}}
& LLaDA-Instruct                 &  1.00  & 7.2  & 0.17$\times$ & 77.4    \\
&               & Dream-Base     &  1.00  & 9.5  & 0.23$\times$  & 75.0   \\
&              & Fast-dLLM (DC)  &  2.10  & 49.8 & 1.19$\times$ & 75.0    \\
&                 & D2F          &  3.10  & 91.2 & 2.18$\times$ & 77.6    \\
%&                 & D2F-LLaDA      & 52.5 & 2.8  & 77.3 \\
\midrule
\multirow{9}{*}{\textbf{MATH}} 
& \multirow{5}{*}{\textit{AR-based}} & AR        &     1.00   &  41.3 &  1.00$\times$ & 77.0    \\
&                                     & Jacobi   &     1.02   &  41.0 &  0.99$\times$ & 77.0     \\
&                                      & CLLM*   &    2.23   &  84.4 & 2.04$\times$ & 77.2     \\
&                                    & \sysname       &   3.82   &  150.7 & 3.65$\times$ & 77.4     \\
&                                  & \sysname (MR)  &  \textbf{3.98} & \textbf{152.0}  &  \textbf{3.68$\times$}  &  77.4    \\
\cmidrule(l){2-7}
& \multirow{4}{*}{\textit{Diffusion-based}} 
&            LLaDA-Instruct     &  1.00  & 21.1 & 0.51$\times$ & 23.7  \\
&            & Dream-Base      &  1.00   & 9.9  & 0.24$\times$  & 35.8    \\
&             & Fast-dLLM (DC)  &  1.90    & 67.0 & 1.62$\times$  & 37.1    \\
&              & D2F           &   2.60    & 98.8 & 2.39$\times$  & 35.4     \\
\bottomrule
\end{tabular}
}%
\label{tab:math_benchmarks}
\end{table}

Moreover, we report the speedup and problem solve rate (test@1) on GSM8K and MATH in Table~\ref{tab:math_benchmarks}.  Across both benchmarks, the Jacobi Forcing Model substantially outperforms the AR baseline with 3.70$\times$ speedup while preserving competitive accuracy. In the MATH benchmark, \sysname delivers a 150.7 TPS while even slightly improving the solve rate from 77.0\% to 77.4\%, highlighting its ability to achieve both high efficiency and accuracy.

We also present speedup comparison across different AR-based techniques with \sysname on B200 in Table~\ref{tab:b200_speedup} as it comes with a better fast-forward count to TPS conversion rate with more compute on B200.

\begin{wraptable}{r}{0.49\textwidth}
\centering
\vspace{-0.5cm}
\small
\setlength{\tabcolsep}{6pt}
\caption{
Speedup on HumanEval tested on B200 using same settings and speedup ratio over A100. %\todo{rerun pcllm over 10-16 ckpt}
}
\vspace{-0.05cm}
\resizebox{\linewidth}{!}{%
\begin{tabular}{l c c c c}
\toprule
 Method & {TPF $\uparrow$} & {TPS $\uparrow$} & {Speedup $\uparrow$} \\
\midrule
AR        &  1.0  &    83.0     &  1.00$\times$     \\
Jacobi    &  1.03  &    84.7     &  1.02$\times$        \\
CLLM*     & 2.68  &    207.4    &     2.49$\times$     \\
\sysname   &  4.01  &    301.7   &  3.63$\times$       \\
\sysname (MR) & 4.21  &  \textbf{328.0}    &  \textbf{3.95$\times$}    \\
\bottomrule
\end{tabular}
}%
\vspace{-0.2cm}
\label{tab:b200_speedup}
\end{wraptable}

On B200, with the block size at 128 and verification size at 4 (rationale provided in Section~\ref{sec:exp_ablation}), we apply multi-block decoding using \sysname and the results are presented in Figure~\ref{fig:mutiblock_and_rejection_recycling}. The running window method is an optimized variant of Jacobi decoding designed for settings where many tokens are accepted per iteration. It maintains a fixed-size active block by replenishing draft tokens to the original block size as accepted tokens are committed to the KV cache. The results demonstrate that multi-block decoding with rejection recycling consistently achieves the highest number of fast-forwarded tokens per iteration, particularly in the larger block-size regime as shown in Figure~\ref{fig:mutiblock_and_rejection_recycling_configuration_search}.

\subsection{Ablation Study}
\label{sec:exp_ablation}

\textbf{Training Noise schedules.} We evaluate three types of noise schedules: random, linear progressive, and reverse progressive. In the random schedule, the noise step $t_i$ for each block is sampled uniformly as $t_i \sim \mathcal{U}(1, \dots, N)$ during sequence packing in \sysname training. The linear progressive schedule follows Eq.~\ref{eq:noise_schedule}, while the reverse progressive schedule applies a linearly decreasing noise ratio from 1 to 0 within each window. Results in Table~\ref{tab:noise_schedule} show that the linear progressive schedule significantly outperforms the other two when the window size is 8. Intuitively, with $N=16$, this schedule corresponds to adding noise more aggressively across blocks within each window, roughly two additional noisy tokens per future block, until the final block where all tokens are noisy.

\textbf{Training Mask types.} We train \sysname on the objective in Eq.~\ref{eq:cllm2_consistency_loss} with noise-conditioned mask implementation (Figure~\ref{fig:noise_aware_causal_mask}). An alternative implementation of the mask is to condition all blocks within a window on a clean context. In other words, for every query, it sees blocks from all preceding windows as of Figure~\ref{fig:clean_context_causal_mask}]), and all blocks within its own window as of Figure~\ref{fig:noise_aware_causal_mask}. Intuitively, it makes token predictions in later windows and blocks easier to learn because now they are conditioned on a cleaner context. We summarize results in Table~\ref{tab:mask_type_ablation}, where it shows noise-conditioned mask is more effective in empowering \sysname with speedup while maintaining generation quality.

%\lx{add mask types: cllm1, cllm2, (and potentially) window-based noise-aware mask}

\begin{table}[t]
\centering
\small
\setlength{\tabcolsep}{6pt}
\vspace{-0.1cm}
\caption{Inference results for block size = 256 with $N=16$, $t_{\text{min}} = 0.0$ and $t_{\text{max}}=1.0$. Acc. = pass@1 accuracy (\%) on HumanEval. The checkpoints are trained with Qwen2.5-Coder-7B-Instruct on 10k randomly sampled instances from our OpenCodeInstruct trajectory dataset. Notice that for ablation purpose, the checkpoints are not trained with full datascale as in Table~\ref{tab:coding_benchmarks}.
%and Table~\ref{tab:math_benchmarks}. 
Reverse progressive is significantly worse than other schedule and we only conduct ablation for one choice of window size.}
%\begin{tabular}{ccccccccc}
%\toprule
%\multirow{2}{*}{Block Size} & \multicolumn{2}{c}{Random} & \multicolumn{2}{c}{Linear Progressive} & \multicolumn{2}{c}{Random Progressive} & \multicolumn{2}{c}{Reverse Progressive} \\
%\cmidrule(lr){2-3} \cmidrule(lr){4-5} \cmidrule(lr){6-7} \cmidrule(lr){8-9}
% & Acc. & iter/token & Acc. & iter/token & Acc. & iter/token & Acc. & iter/token \\
%\midrule
%8  & 82.9 & 0.53 & 84.7 & 0.48 & 82.3 & 0.48 & --   & --   \\
%16 & 83.5 & 0.51 & 81.7 & \textbf{0.46} & --   & --   & 82.9 & 0.62 \\
%\bottomrule
%\end{tabular}
%\label{tab:noise_schedule}
%\end{table}

\begin{tabular}{ccccccc}
\toprule
\multirow{2}{*}{Window Size} & \multicolumn{2}{c}{Random} & \multicolumn{2}{c}{Linear Progressive}  & \multicolumn{2}{c}{Reverse Progressive} \\
\cmidrule(lr){2-3} \cmidrule(lr){4-5} \cmidrule(lr){6-7}
 & Acc. & iter/token & Acc. & iter/token & Acc. & iter/token \\
\midrule
8  & 82.9 & 0.53 & \textbf{84.7} & 0.48    & --   & --   \\
16 & 83.5 & 0.51 & 81.7 & \textbf{0.46}   & 82.9 & 0.62 \\
32 & 83.5 & 0.53 & 84.1 & 0.49 & -- & --\\
\bottomrule
\end{tabular}
\label{tab:noise_schedule}
\end{table}

%\todo{maybe (1) training data size, (2) progressive distillation performance vs. effiicency tradeoff.}

\begin{wraptable}{r}{0.49\textwidth}
\centering
\vspace{-0.2cm}
\small
\setlength{\tabcolsep}{6pt}
\caption{
Effects of applying noise-conditioned mask (NC) or noise-conditioned mask with intra-window clean context (NC-IC) for \sysname training, and evaluated on HumanEval with A100.
}
\vspace{-0.05cm}
\begin{tabular}{l c c c}
\toprule
 Method &  {Speedup$\uparrow$} & Acc. \\
\midrule
NC            &  \textbf{3.6$\times$}    &   \textbf{82.3} \\
NC-IC          & 1.9$\times$     &  82.3   \\
\bottomrule
\end{tabular}
\label{tab:mask_type_ablation}
\end{wraptable}

\begin{figure}[t]
    \centering
    % -------- left image ----------------------------------------
    \begin{minipage}[l]{0.50\linewidth}
        \centering
        \includegraphics[width=0.9\linewidth]{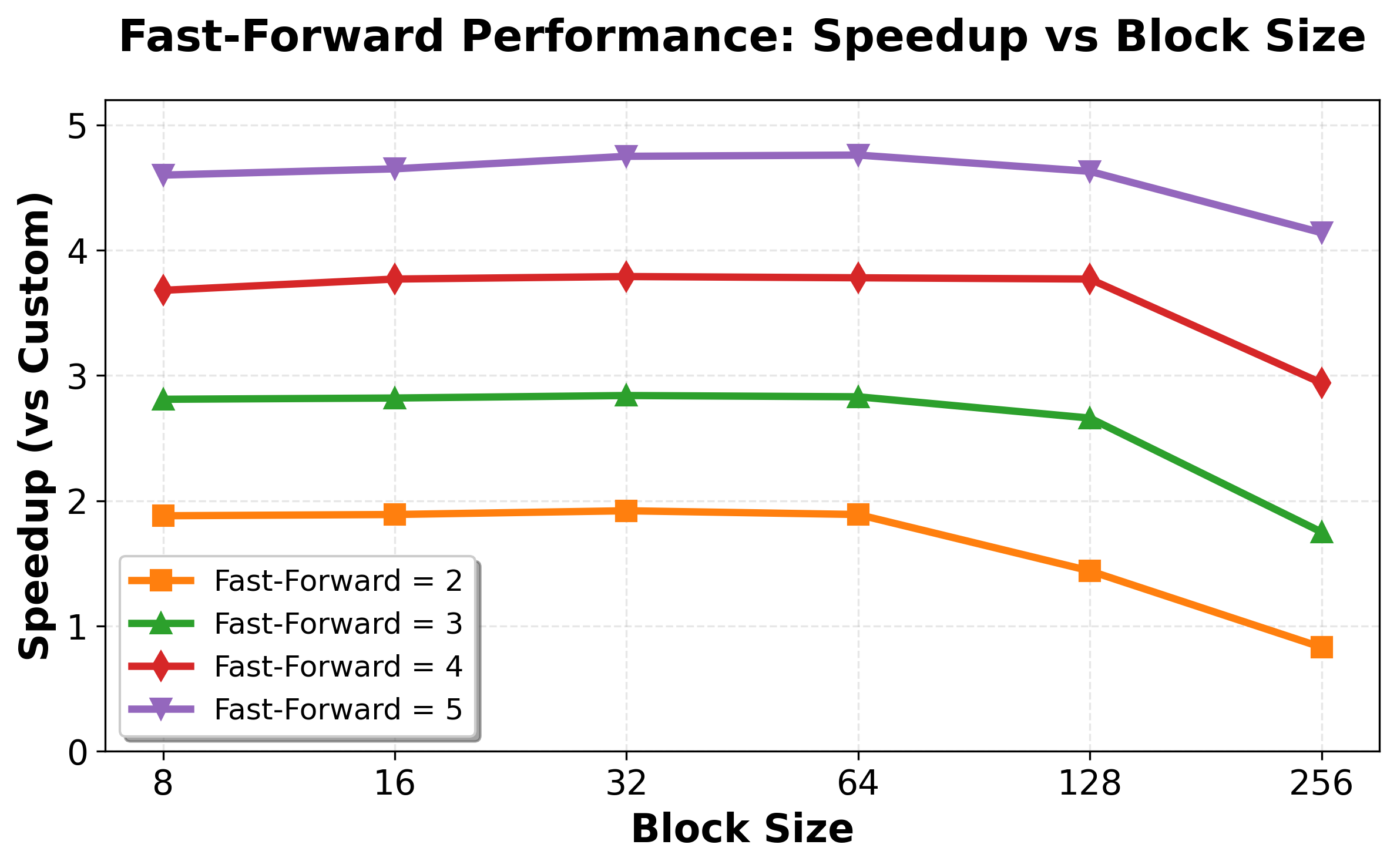}
        \subcaption{Speedup vs. (log-scaled) block size at fixed fast-forwarding count per iteration on NVIDIA H200 GPU, using Jacobi decoding at prompt length = 128, generation length = 256 at varying TPF rates.}
         \label{fig:ffcount}
    \end{minipage}
    % -------- right image ---------------------------------------
    \begin{minipage}[c]{0.425\linewidth}
        \vspace{-0.05cm}
        \centering
        \includegraphics[width=\linewidth]{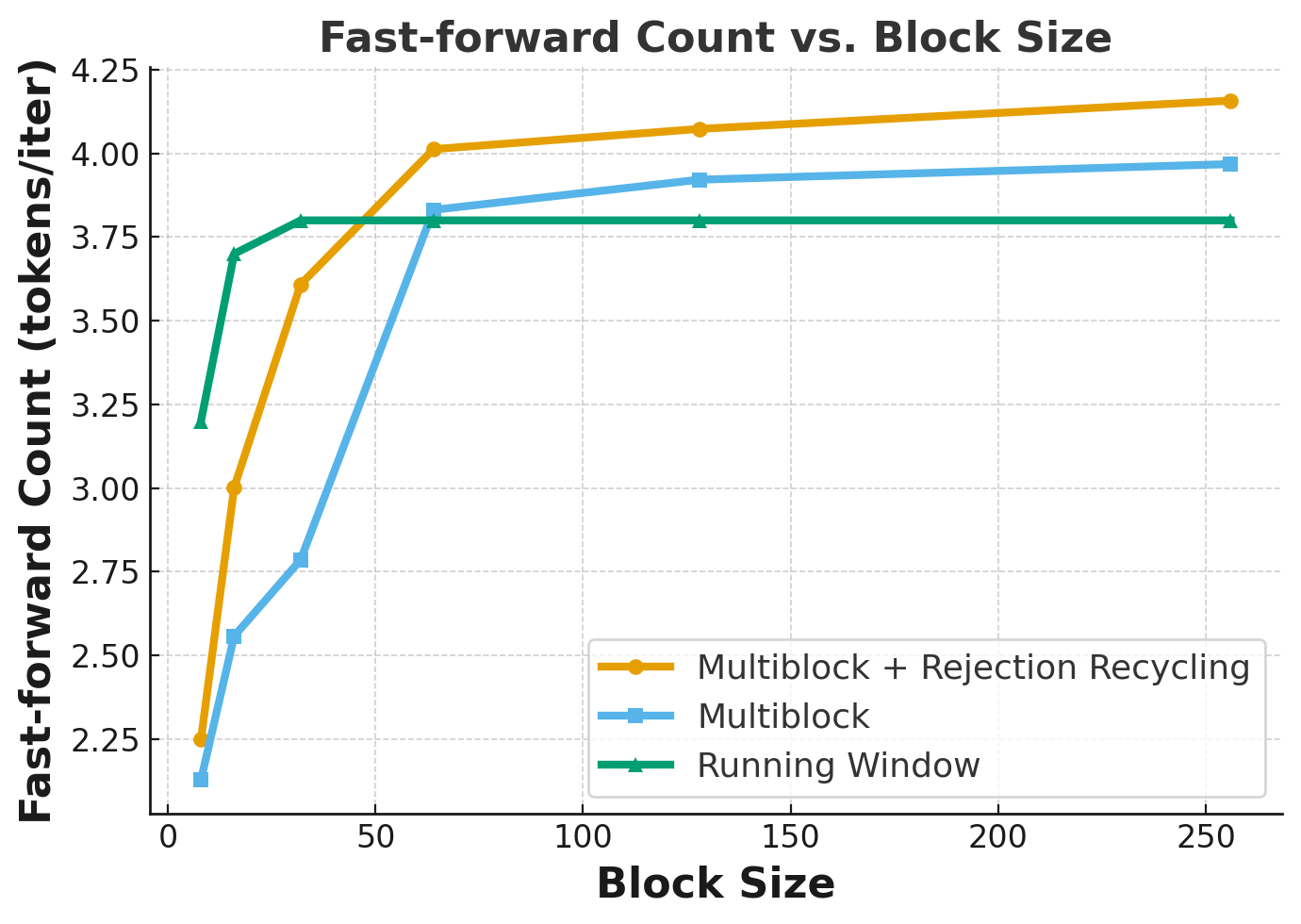}
        \subcaption{fast-forward count vs. block size on HumanEval using three decoding strategies on NVIDIA H200 GPU. Notice larger block size provides more fast-forward token count for multi-block decoding with rejection recycling.}
        \label{fig:mutiblock_and_rejection_recycling_configuration_search}
    \end{minipage}
    \hfill
    % ------------------------------------------------------------
    \caption{Effect of block size choices on fast-forward counts and wall-clock speedup under different settings. We choose the maximum block size on hardware without sacrificing wall-clock speedup. 
    %\todo{replace with latest profiling figures}
    }
    \vspace{-0.5cm}
    \label{fig:blocksize_and_inference}
\end{figure}

\textbf{Inference FLOPs Utilization Analysis.} \sysname (MR) involves both multi-block decoding and rejection-recycling, where each technique consumes extra FLOPs for parallel drafting and parallel verification, respectively. To maximize hardware utilization, we experiment with how end-to-end decoding latency changes as the total number of decoded tokens changes. We use Jacobi decoding to run the experiments and the results are shown in Figure~\ref{fig:ffcount}. On H200 GPUs, Jacobi decoding with block sizes up to 64 shows no latency penalty and only minor degradation at 128, particularly in the high fast-forwarding regime. The result is consistent across accepted token counts fixed at $2,3,4,5$, indicating that up to 126 tokens can be decoded in parallel with shared KV without significant latency overhead. We provide a more detailed analysis in Appendix~\ref{appendix:tps_and_flops_tradeoff}.

%\begin{wrapfigure}{l}{0.49\textwidth}
%  \centering
%    \includegraphics[width=\linewidth]{figures/fast_forward_performance.png}
%    \caption{Speedup vs. block size under different fast-forwarding count per iteration on NVIDIA H200 GPU.}
%    \label{fig:ffcount}
%\end{wrapfigure}

%\begin{wrapfigure}{r}{0.49\textwidth}
%  \centering
%  \includegraphics[width=\linewidth]{figures/fast_forward_vs_blocksize.png}
%    \caption{Multiblock decoding with rejection recycling.}
%\label{fig:mutiblock_and_rejection_recycling}
%\end{wrapfigure}

%In this section, we analyze how block size impacts speedup for vanilla Jacobi decoding and multiblock decoding using \sysname. As discussed in Section~\ref{sec:exp_results}, modern GPUs can sustain large block sizes without increasing end-to-end latency compared to the AR baseline.

\textbf{Inference Configuration Search.} Beyond block size, the main tunable parameters for \sysname (MR) inference are verification size (entries verified in parallel with shared KV for rejection recycling), number of blocks, and initialization threshold. We observe that performance gains from additional blocks saturate at block size = 2 as later drafts degrade quickly. The initialization threshold, defined as the fraction of the first block completed before launching the next, can be optimized via grid search and shows consistently optimal performance at $r=0.85$ for block size 64 across verification sizes 2 to 8. For maximum FLOPs utilization, we use block size = 64, verification size = 4, where wall-clock speedup remains stable until parallel decoding exceeds 256 tokens. More details on inference configuration search given the FLOPs budget can be found in Appendix~\ref{appendix:inference_configuration_grid_search}.

% Once $K=2$ and $r=0.85$ are fixed as training-optimal values from a separate grid search, the remaining degrees of freedom at inference are the block size $n_{\text{token\_seq\_len}}$ and the $n$-gram verification size, which jointly determine how much parallel draft and verify work is done per step under a given hardware constraint. To explore this space, we perform a grid search over block sizes
% $n_{\text{token\_seq\_len}} \in \{8,16,32,64,128,256\}$ and $n$-gram verification sizes $n_{\text{gram}} \in \{1,2,4,8,12\}$, measuring the achieved tokens per second for each pair on the target GPU. The results are shown in Figure~\ref{fig:block_ngram_tps}, and the resulting surfaces reveal a clear optimum region: tokens-per-second peaks at moderate block sizes and medium $n$-gram verification, with the global maximum near $n_{\text{token\_seq\_len}}\approx 64$ and $n_{\text{gram}}\approx 4$. Very small blocks or $n$-gram verification size underutilize the available FLOPs, while very larger choices push the system closer to the roofline and begin to degrade wall-clock latency. This analysis justifies the final choice of using block size $64$ and $n$-gram size $4$ on H200/B200, which lies near the empirical optimum under each GPU’s FLOPs budget. 

% \input{figure_text/inference_grid_search}
%\lx{add wall-clock speedup vs. block size and verification size plots.}

%\lx{add a FLOPs scaling experiments.}

\section{Related Work}

\textbf{Discrete Text Diffusion.} dLLMs represent a new paradigm that challenges traditional autoregressive (AR) modeling by replacing left-to-right causality with iterative denoising, enabling parallel multi-token generation~\citep{li2024derivativefree, nisonoff2024unlocking, schiff2024simpleguidance}. Closed-source dLLMs (e.g., Gemini Diffusion~\citep{gemini_diffusion_2025, khanna2025mercury, seed_diffusion_2025}) show huge throughput improvement while maintaining competitive code and text quality, underscoring better accelerator utilization. On the open-source side, community dLLMs with released code and weights delivered strong throughput and controllability via parallel iterative denoising, yet remaining less efficient than autoregressive decoding~\citep{ye2025dream7b, zhu2025llada15_vrpo, nie2025large_language_diffusion_models, jetastraSDAR, gong2024scaling_diffusion_language_models}. Recent efforts~\citep{arriola2025block, wu2025fast_dllm, liu2025dllm_cache} further push the efficiency and scalability of dLLMs.

\textbf{Jacobi Decoding.} Jacobi decoding reframes AR generation as a parallel fixed-point update over all positions, with convergence linked to greedy AR, and has been instantiated using Jacobi (Gauss-Seidel) iterations~\citep{song2021accelerating,santilli2023accelerating}. Building on this, follow-ups either refine the decoding procedure or train models as parallel decoders to exploit parallel: CLLMs~\citep{kou2024cllms_consistency_large_language_models} fine-tune LLMs with consistency distillation to predict multiple correct tokens per iteration and speed convergence; CEED-VLA~\citep{song2025ceed_vla_consistency_vla} brings the similar idea to robotics. Other strands adapt Jacobi to new regimes, including FastCoT~\citep{zhang2023fast} for reasoning with parallel CoT updates, Speculative Jacobi Decoding~\citep{teng2024accelerating} for sampling in AR Test-to-Image, and MSN, TR-Jacobi~\citep{wang2024make} that injects denoising training and a retrieval-augmented Jacobi strategy.

\textbf{Speculative Decoding}. Speculative decoding speeds up AR generation by letting a lightweight drafter propose several future tokens and having the target model verify them in one pass~\citep{leviathan2022speculative_decoding,chen2023accelerating}. It preserves the target model's distribution while reducing latency. Subsequent work improves proposal quality and verification efficiency: online speculative decoding (OSD)~\citep{liu2024online_speculative_decoding} adapts draft models to user query distributions via continual distillation, substantially improving token acceptance and reducing inference latency. Medusa~\citep{cai2024medusa} adds multi-head drafters to the base LM to produce verifiable token blocks; EAGLE, EAGLE-2~\citep{li2024eagle,li2024eagle2} reuse target features for feature-level drafting, and EAGLE-3~\citep{li2025eagle3} scales this idea with multi-layer fusion. Lookahead Decoding~\citep{fu2024lookahead}, PLD~\citep{saxena2023pld,somasundaram2024pld}, and REST~\citep{he2023rest} dispense with a separate drafter, instead synthesizing speculative candidates directly from context or future tokens. The self-speculative decoding paradigm shares a close connection with the Jacobi decoding adopted in this work.

\section{Conclusion}

%In this work, we introduce a progressive distillation technique to train AR models as faster and more accurate parallel decoders than dLLMs. Specifically, instead of teaching model to learning predicting a large block of tokens in parallel like CLLM does, our method involves teaching the AR models to learn progressively difficult leaning objective by applying a progressive noise schedule with a new sequence packing technique and noise-aware causal mask to train parallel token predictions conditioning on noise. We also train the model iteratively by regenerating trajectories with progressively large block sizes. The model trained this way, \sysname, shows significant speedup at around $3.6\times$ while largely maintaining model accuracy. Upon inspecting trajectories generated by the model trained this way, \sysname, exhibit high-quality draft tokens at the trails. We further introduce rejection recycling and multi-block decoding, which brings model TPS to up to $4\times$ on H200 GPU on HumanEval. 

In this work, we propose a progressive distillation technique for training AR models as faster and more accurate parallel decoders compared to dLLMs. Unlike CLLM~\citep{kou2024cllms_consistency_large_language_models}, which directly trains models to predict large blocks of tokens in parallel, our approach introduces a progressively more difficult learning objective. This is achieved through a progressive noise schedule, combined with a sequence packing strategy and a noise-aware causal mask, enabling parallel token prediction conditioned on noise. The model is further improved through iterative training, where trajectories are regenerated with progressively larger block sizes. The resulting model, \sysname, achieves a 3.8$\times$ speedup while largely preserving accuracy. Analysis of its generated trajectories shows that \sysname produces high-quality draft tokens toward the tail of sequences. In addition, we introduce rejection recycling and multi-block decoding, which together bring tokens accepted per iteration to $4.5\times$ as high with nearly $4\times$ speedup on HumanEval using on both A100 and B200 GPUs.

\newpage
\section*{Ethics Statement}

All authors have read and adhere to the \href{https://iclr.cc/public/CodeOfEthics}{ICLR Code of Ethics}. This work does not involve human subjects, sensitive personal data, or experiments with the potential to cause harm. No confidential or proprietary data were used. The methods and experiments were conducted in accordance with principles of research integrity, fairness, and transparency. Potential societal impacts, including limitations and biases of large language models, are explicitly discussed in the paper. All conclusions are the sole responsibility of the authors.

\section*{Reproducibility Statement}

We have made significant efforts to ensure the reproducibility of our results. Detailed descriptions of the models, datasets been used, as well as hyperparameter choices are included in the main text. All datasets used are publicly available, and the preprocessing steps are fully documented. Ablation studies are provided to validate robustness of results. These resources collectively allow independent researchers to verify and reproduce our work.

\section*{Use of LLM}

During the preparation of this manuscript, large language model was used to refine grammar and improve clarity. The authors carefully reviewed and revised all outputs to ensure the text reflects their original ideas and take full responsibility for the final content, including all statements and conclusions.

\bibliography{cllm2}
\bibliographystyle{iclr2026_conference}

\newpage
\appendix

\section{Detailed Decoding Algorithm}
\begin{algorithm}[t]
\caption{\textsc{MultiBlock Decoding + Rejection Recycling}}
\label{alg:multiblock}
\begin{algorithmic}[1]
\State \textbf{Init:} Create a set of blocks $\{b\}$ with one \emph{real–active} block $RA$: draft tokens $q_{RA}$ randomly initialized, accepted tokens $a_{RA}=\varnothing$
%, counter $t_{RA}=0$
; For all other blocks $b$, set $q_b=\varnothing$, $a_b=\varnothing$, 
%$t_b=0$, 
and mark as \emph{pseudo-active}. \\
Initialize candidate pool $\mathcal{N}=\emptyset$, spawn ratio $r$, threshold $s=\lceil r n\rceil$, block size $n$.
\While{iters $<$ max}
  \State \textbf{Assemble input $\vy$:} Concatenate $q_{RA}$, then for each pseudo-active $b$, append $a_b$ (no logits) and $q_b$ (collect logits). Resize cache to batch $\vy$.
  \State \textbf{Forward:} Run model $p_{\theta}(\vy)$ to produce logits.
  \For{each block $b$ with span $(start,L)$}
    \State \textbf{Verification (with rejection-recycling):} Greedy prediction $g=\arg\max$ logits; accept longest matching prefix of $q_b$ using $g$ (or $g \; \cup \; \mathcal N$ if $b=RA$); update $a_b$.
    \If{$b=RA$ and EOS encountered in accepted region} \State \Return committed output. \EndIf
    \State \textbf{Tail update:} If partial accept, set $q_b \gets [\text{next} \Vert g_{\text{tail}}]$ (and if $b=RA$: push rejected tail to update $\mathcal{N}$ and $q_{RA}$); else $q_b \gets \varnothing$.
  \EndFor
  \State \textbf{Cache trim:} Delete false KV to committed length: prompt $+$ verified $a_b$ (all accepted blocks) $+$ $a_{RA}$.
  \State \textbf{Spawn:} If some block $b$ reaches $\vert a_b \vert \ge s$ and active $\left\{ b \right\} <K$, clone and pad $q_{RA}$ to length $n$ and add as new pseudo-active block.
  \State \textbf{Promote:} If $\vert a_{RA} \vert \ge n$, choose a pseudo-active $b$ with $\vert a_b \vert >0$, rebuild its draft to length $n$, mark as verified, set $RA \gets b$.
  \State \textbf{Stop:} If all $\vert a_b \vert \ge n$ or EOS emitted by $RA$, break.
\EndWhile
\State \textbf{Finalize:} Concatenate $\text{output } =$ verified $a_b$ for all non-RA blocks, then $a_{RA}$; trim KV cache $\mathcal{C}$; 
\State \textbf{Return:} $(\text{output}, \mathcal{C}, \text{iters})$
\end{algorithmic}
\end{algorithm}

\label{appendix:algorithm}
We present the detailed algorithm for multi-block decoding and rejection sampling introduced in Section~\ref{sec:cllm2_inference}. Rejection recycling reuses high-quality consecutive tokens discarded in previous Jacobi iterations to construct candidate token sequences. Multi-block decoding jointly maintains and refines multiple blocks, allowing correct tokens in later blocks to be decoded even when earlier blocks remain unconverged, thereby further improving decoding throughput. These two techniques are orthogonal and can be seamlessly combined. As shown in Table~\ref{tab:b200_speedup}, their combination yields an improvement of over 30 TPS compared to vanilla Jacobi decoding on a B200 GPU.

\section{Further Baseline Comparisons}
\label{appendix:additional_comparisons}

The main text focuses on comparisons between \sysname\ and diffusion-based
parallel decoders, as well as AR-based parallel decoders, under a controlled
setup where AR variants share the same backbone (Qwen2.5-Coder-7B-Instruct).
This appendix extends the comparison to (i) distilled discrete diffusion
models and (ii) state-of-the-art speculative decoding baselines.

\begin{table}[htb]
\centering
\small
\setlength{\tabcolsep}{6pt}
\caption{
Additional comparison on HumanEval across AR, speculative decoding,
and dLLM-based methods. For the AR baseline and all Jacobi-decoding based methods, Qwen2.5-Coder-7B-Instruct is used as the backbone. Speedup is measured in TPS relative to the AR baseline on a single B200 GPU.
}
\label{tab:specdec_dllm}
\resizebox{\columnwidth}{!}{%
\begin{tabular}{llcccc}
\toprule
Family              & Method          & Acc. $\uparrow$ & TPF $\uparrow$ & TPS $\uparrow$ & Speedup vs.\ AR $\uparrow$ \\
\midrule
AR             & AR (greedy)     & 87.8           & $1.00$         & $83.00$        & $1.00\times$                \\
dLLM                & Fast-dLLM v2    & 63.4           & $1.00$         & $83.29$        & $1.00\times$                \\
dLLM                & SDAR            & 78.7           & $2.36$         & $31.46$        & $0.38\times$                \\
dLLM (distilled)    & dParallel       & 54.3           & $2.90$         & $175.15$       & $2.11\times$                \\
\midrule
AR + Spec-Dec       & EAGLE-3$^{*}$   & 68.9$^{*}$     & $6.38$         & $246.10$       & $2.97\times$                \\
AR + Spec-Dec       & HASS$^{*}$      & 61.6$^{*}$     & $5.53$         & $280.29$       & $3.37\times$                \\
\midrule
AR + Jacobi         & Jacobi          & 87.8          & $1.05$         & $84.70$        & $1.02\times$                \\
AR + Jacobi       & CLLM            & 87.8           & $2.68$         & $207.40$       & $2.50\times$                \\
AR + Jacobi         & \sysname           & 83.5           & $4.01$         & $301.65$       & $3.63\times$                \\
AR + Jacobi         & \sysname (MR)     & 83.5           & 4.21         & \textbf{327.96}       & $\mathbf{3.95\times}$                \\
\bottomrule
\end{tabular}%
}
\vspace{0.2em}
\vspace{0.2em}
\footnotesize{
\parbox{\columnwidth}{
$^{*}$Here we report the strongest checkpoints released by the authors, in principle EAGLE-3 and HASS are lossless in comparison with greedy AR checkpoints if they were trained with the Qwen2.5-7B backbone.
%\todo{rerun with 10-16 ckpts for JaF-LLM}
}}
\end{table}

\paragraph{Distilled dLLM baselines.}
A distilled dLLM baseline is useful for mapping \sysname against contemporary
training techniques for discrete diffusion models. dParallel~\citep{chen2025dparallel} performs trajectory-level consistency distillation on a discrete diffusion model to accelerate token sampling while aiming to preserve quality. We adopt the technique as the latest distilled dLLM baseline.

As shown in Table~\ref{tab:specdec_dllm}, on HumanEval, \sysname (MR) attains a noticeably stronger speed–quality profile
than dParallel: \sysname (MR) achieves $29\%$ higher accuracy and
achieves more than $80\%$ higher TPF and TPS. On GSM8K, \sysname improves accuracy by $8$ absolute points with about $20\%$ higher TPF and TPS (GSM8K numbers are omitted from the table below for brevity). These gaps indicate that, relative to latest consistency-distilled dLLM of comparable scale, \sysname occupies a more favorable point in the speed–quality trade-off space.

\paragraph{Speculative decoding and recent dLLM baselines.}
Speculative decoding (SD) forms widely used family of AR
acceleration methods. To place \sysname\ among such approaches, this appendix
includes comparisons against two recent SD methods, EAGLE-3~\citep{li2025eagle3}
and HASS~\citep{zhang2025hass}, which represent stronger baselines than earlier
methods such as Medusa and Medusa-2.

The comparison in Table~\ref{tab:specdec_dllm} also includes two recent dLLM baselines, Fast-dLLM v2~\citep{wu2025fastdllmv2} and SDAR~\citep{cheng2025sdar}, in addition to the community dLLM and D2F variants discussed in the main text. Fast-dLLM v2 improves blockwise diffusion efficiency via enhanced scheduling and caching, while SDAR introduces a synergistic diffusion–autoregressive paradigm for scalable sequence generation.

\section{Mapping Noise Schedule to Training Sequence for Progressive Consistency Distillation}

\begin{figure}[tbh]
    \centering
    \vspace{-0.2cm}
    \includegraphics[width=1.0\linewidth]{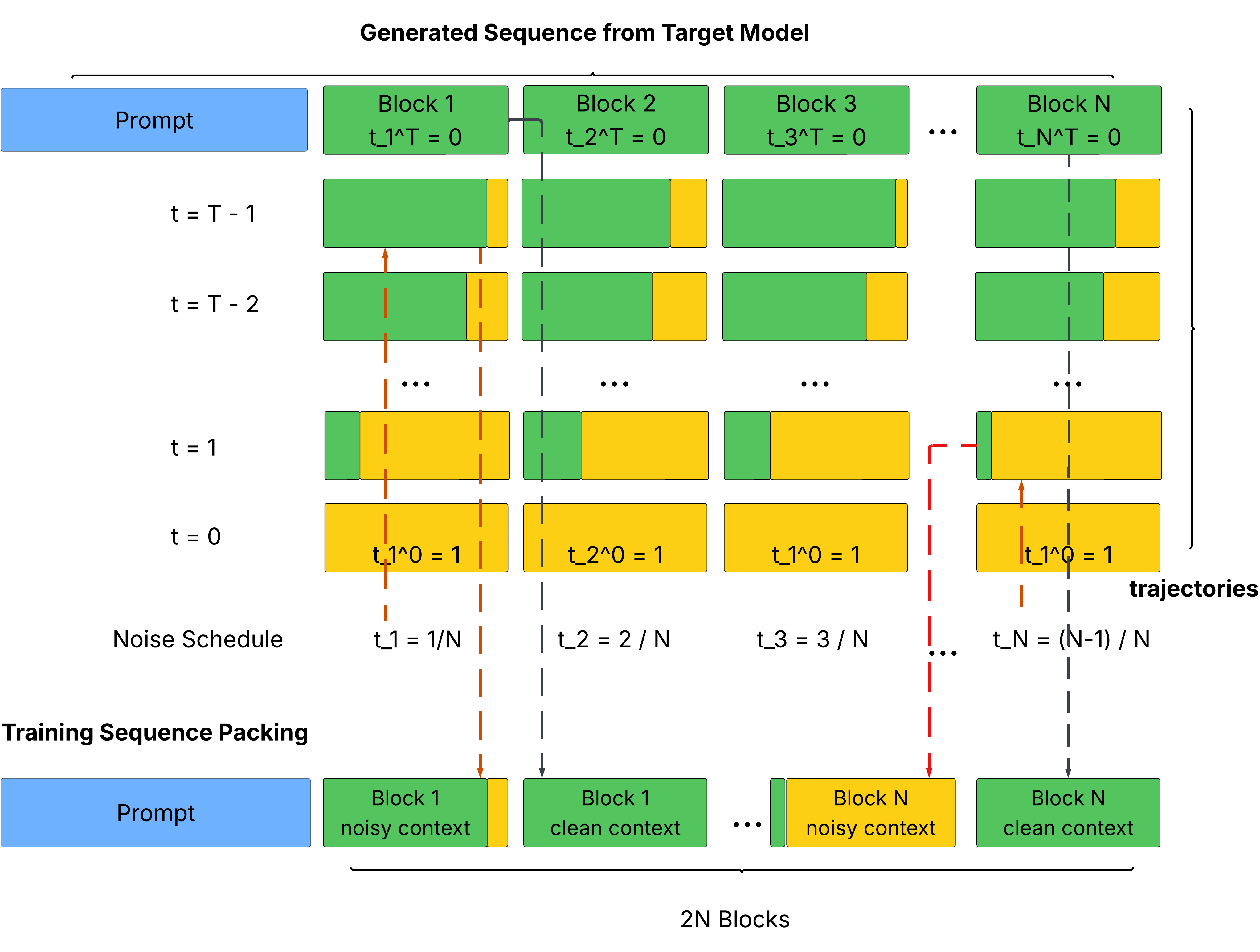}
    \caption{
Illustration of the progressive noise schedule and training sequence packing. For each block $i$ over a total of $T_i$ decoding steps, we select the trajectory step whose fraction of unconverged tokens matches the scheduled noise ratio $t_i$ to form a noisy block (dashed red line), and pair it with the corresponding clean block (dashed dark line). The packed training sequence at the bottom interleaves all noisy and clean blocks, yielding $2N$ blocks so that a single forward pass can compute both AR and consistency losses.
}
\label{fig:noise_schedule_and_training_seq_mapping}
\end{figure}

We elaborate the process of mapping the noise schedule to arrive at the training sequence in Figure~\ref{fig:sequence_packing_and_mask}.

For each training sample, let the target model’s complete generation of length
$L$ be $\vy$. Given a training-time block size $n$ and a noise schedule $W$
(e.g., the linear progressive schedule in Eq.~2), we partition $\vy$ into
$N = \lceil L / n \rceil$ blocks of size $n$. The schedule $W$ is applied over
a window of $w$ blocks, yielding noise ratios $t_i$ defined in Eq.~\ref{eq:noise_schedule}.
For each block, we select the point along its Jacobi trajectory whose fraction of unconverged tokens $(\text{number of unconverged tokens}/n)$ is closest to $t_i$, and use that point to form the corresponding noisy block. A full illustration is shown in Figure~\ref{fig:noise_schedule_and_training_seq_mapping}.

A complete training sequence contains both noisy and clean blocks. Clean blocks are the original partitions of $\vy$, while noisy blocks are constructed as above. We interleave each noisy block with its corresponding clean block so that a single forward pass, together with the custom attention mask in Figure~\ref{fig:trajectory_vis}, produces teacher logits on clean blocks for the AR loss and student logits on noisy blocks for the consistency loss. Under the progressive noise schedule, the longest consecutive noisy span within any block is $O(\lceil t n\rceil)$, which is much smaller than the naive $O(nN)$ worst case where every token in every block is noisy.

\section{Understanding TPF and FLOPs Trade-off}
\label{appendix:tps_and_flops_tradeoff}

%\todo{add latest profiling results and visualizations on A100, B200.}

\begin{figure}[t]
    \centering
    \vspace{-0.2cm}
    \begin{subfigure}{0.32\linewidth}
        \centering
        \includegraphics[width=\linewidth]{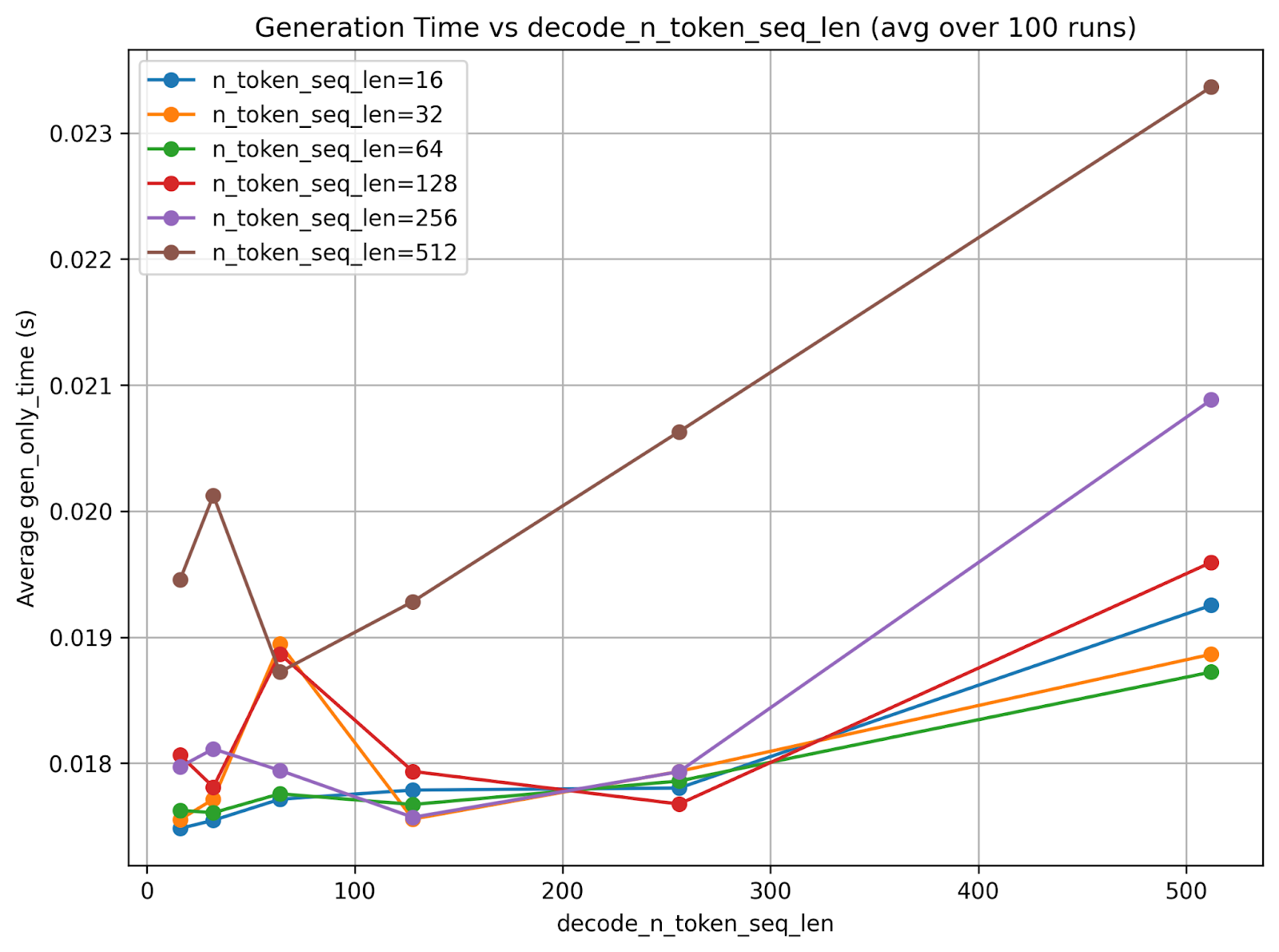}
        \caption{H200}
        \label{fig:h200_profiling}
    \end{subfigure}
    \hfill
    \begin{subfigure}{0.32\linewidth}
        \centering
        \includegraphics[width=\linewidth]{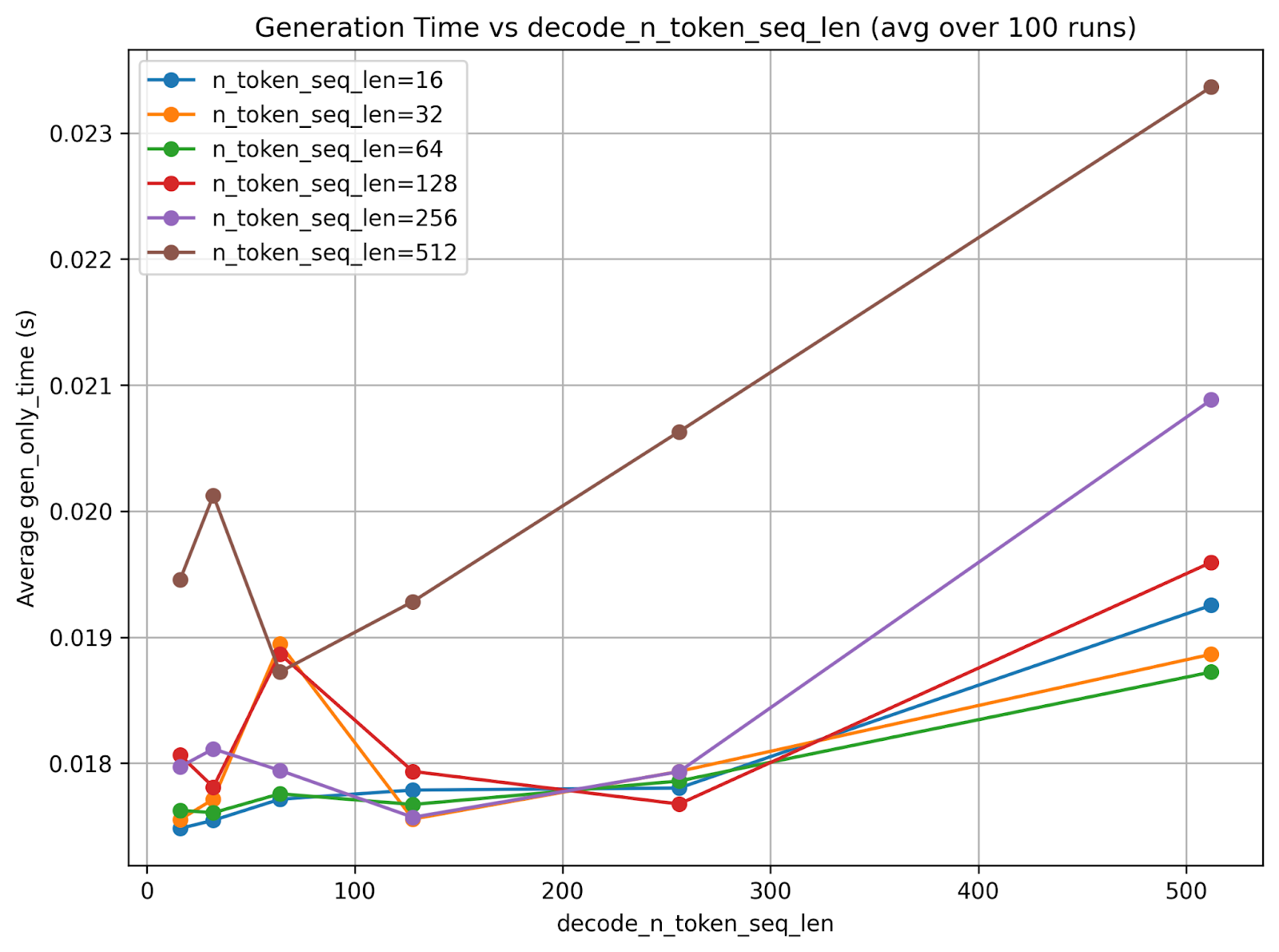}
        \caption{B200}
        \label{fig:b200_profiling}
    \end{subfigure}
    \hfill
    \begin{subfigure}{0.32\linewidth}
        \centering
        \includegraphics[width=\linewidth]{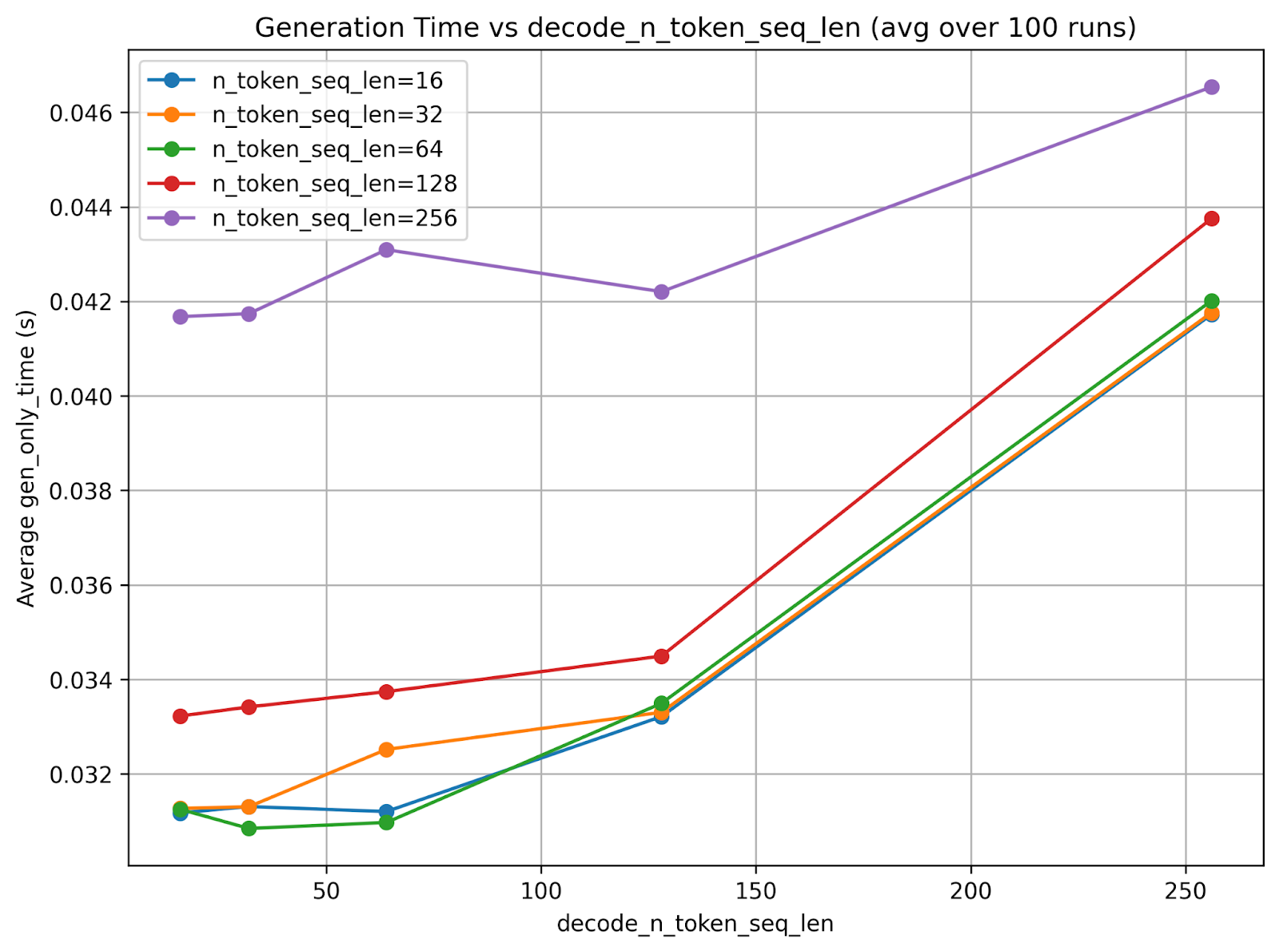}
        \caption{A100}
        \label{fig:a100_profiling}
    \end{subfigure}
    \caption{
Generation-only latency versus total number of parallel decoded tokens across three hardware platforms (A100, H200, B200).
}
    \label{fig:roofline_profiling}
\end{figure}

To estimate how many tokens can be decoded in parallel before hitting the
hardware roofline, we profile generation-only latency as a function of the
total number of simultaneously decoded tokens (horizontal axis in
Figure~\ref{fig:roofline_profiling}), sweeping several block sizes
$n_{\text{token\_seq\_len}}$. On H200 and B200 (left and middle panels), the
curves for $n_{\text{token\_seq\_len}} \in \{16,32,64,128\}$ are essentially
flat as we increase the parallel token count up to $\approx 256$ tokens, and
only start to grow noticeably when we push beyond that to $512$ tokens. This
plateau followed by an approximately linear region is the empirical roofline:
up to $\sim 256$ batched tokens the GPU has spare FLOPs and KV bandwidth, so
extra tokens are almost ``free,'' whereas beyond that point the device becomes
compute- or memory-bound and latency scales roughly linearly.

On A100 (right panel of Figure~\ref{fig:roofline_profiling}), the plateau is shorter: generation time is nearly constant up to $\sim 128$ parallel tokens, but increases steeply once we go beyond 128 and approaches linear scaling by
256 tokens. Taken together, these measurements suggest operating near the ``knee'' of each roofline, which corresponds to $\approx 128$ parallel tokens on A100 and $\approx 256$ parallel tokens on H200/B200. This motivates our final configuration: block size $64$ with verification size $4$ on H200 and B200 ($64 \times 4 = 256$ tokens), which maximizes FLOPs utilization without hurting wall-clock performance.

These roofline measurements imply a FLOPs budget on each GPU: once the parallel
token count approaches the hardware knee, additional tokens incur an almost
linear increase in cost. Consequently, there is an explicit TPF--FLOPs tradeoff: configurations with larger blocks and more aggressive parallelism achieve higher TPF, but the extra FLOPs consumption can saturate the hardware and even degrade wall-clock latency.

% Requires: \usepackage{graphicx} and \usepackage{subcaption}

% Within this budget, allocating more FLOPs to parallel drafting and verification increases TPF (tokens-per-forward) but also pushes the system closer to the roofline. 

\section{Inference Configuration Search}
\label{appendix:inference_configuration_grid_search}

Because of this TPF-FLOPs trade-off, choosing an inference configuration is no longer a matter of simply maximizing block size or verification depth: \textbf{the configuration must respect the FLOPs budget implied by the roofline of the target GPU}. Once $K=2$ and $r=0.85$ (initialization threshold) are fixed as training-optimal values from a separate grid search (as discussed in Section~\ref{sec:exp_ablation}, the remaining degrees of freedom at inference are the block size $n_{\text{token\_seq\_len}}$ and the $n$-gram verification size, which jointly determine how much parallel draft and verify work is done per step under a given
hardware constraint.

To explore this space, we perform a grid search over block sizes
$n_{\text{token\_seq\_len}} \in \{8,16,32,64,128,256\}$ and $n$-gram verification sizes $n_{\text{gram}} \in \{1,2,4,8,12\}$, measuring the achieved tokens per second for each pair on the target GPU. Since the raw grid is relatively coarse, we fit a smooth surface over the discrete measurements and use it as a surrogate for continuous hyperparameter selection. Specifically, we construct a 2D polynomial design matrix in $(\text{block size}, \text{$n$-gram size})$ of total degree up to $6$, select the best degree by mean squared error, and then interpolate the fitted surface onto a dense grid using \texttt{scipy.interpolate.griddata} with a light Gaussian-like smoothing pass.

% Requires \usepackage{graphicx} and \usepackage{subcaption}
\begin{figure}[t]
    \centering
    \begin{subfigure}{0.48\linewidth}
        \centering
        \includegraphics[width=\linewidth]{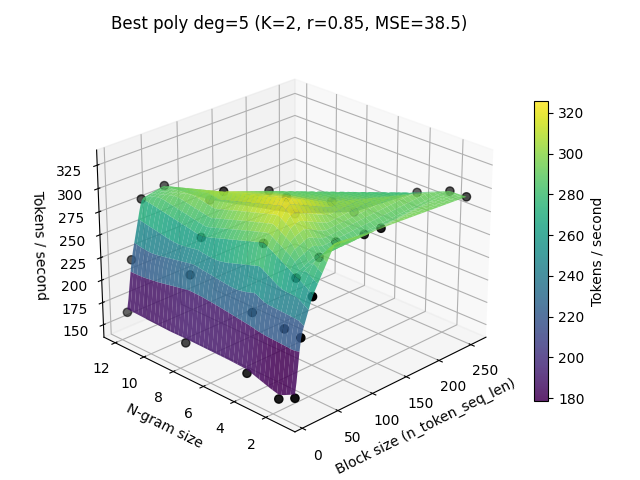}
        \caption{Best-Fit Surface with Interpolation.}
        \label{fig:best_surface_k2_r085}
    \end{subfigure}
    \hfill
    \begin{subfigure}{0.48\linewidth}
        \centering
        \includegraphics[width=\linewidth]{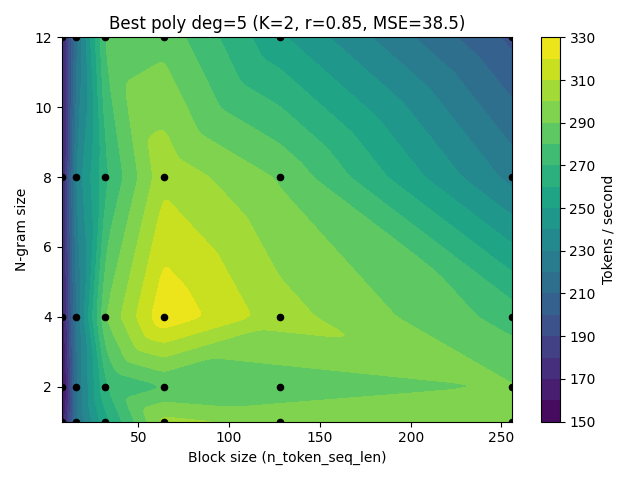}
        \caption{TPS Contour Map.}
        \label{fig:best_contour_k2_r085}
    \end{subfigure}
    \caption{
Tokens-per-second (TPS) as a function of block size $n_{\text{token\_seq\_len}}$ and $n$-gram verification size for $K=2$ and $r=0.85$. Black dots indicate measured configurations; the surface and contours are obtained by Gaussian-like Smoothing.
}
\label{fig:block_ngram_tps}
\end{figure}

The results are shown in Figure~\ref{fig:block_ngram_tps}, and the resulting surfaces reveal a clear optimum region: tokens-per-second peaks at moderate block sizes and medium $n$-gram verification, with the global maximum near $n_{\text{token\_seq\_len}}\approx 64$ and $n_{\text{gram}}\approx 4$. Very small blocks or $n$-gram verification size underutilize the available FLOPs, while very larger choices push the system closer to the roofline and begin to degrade wall-clock latency. This analysis justifies the final choice of using block size $64$ and $n$-gram size $4$ on B200, which lies near the empirical optimum under each GPU’s FLOPs budget.

\end{document}